\documentclass{article} %
\usepackage{iclr2025_conference,times}

\usepackage{amsmath,amsfonts,bm}

\def\eqref#1{equation~\ref{#1}}

\def\1{\bm{1}}

\def\mA{{\bm{A}}}

\def\mD{{\bm{D}}}

\def\mL{{\bm{L}}}
\def\mM{{\bm{M}}}

\def\mS{{\bm{S}}}

\def\mU{{\bm{U}}}
\def\mV{{\bm{V}}}
\def\mW{{\bm{W}}}
\def\mX{{\bm{X}}}
\def\mY{{\bm{Y}}}

\DeclareMathAlphabet{\mathsfit}{\encodingdefault}{\sfdefault}{m}{sl}
\SetMathAlphabet{\mathsfit}{bold}{\encodingdefault}{\sfdefault}{bx}{n}

\usepackage{hyperref}
\usepackage{algorithm}
\usepackage{tikz}
\usepackage{wrapfig}
\usepackage{pgfplots}
\usepackage{graphicx}
\usepackage{subcaption}
\usepackage{adjustbox}
\usepgfplotslibrary{groupplots}
\pgfplotsset{compat=1.17}
\usepackage{algorithmic}
\usepackage{makecell}
\usepackage{multirow}
\usepackage{float}
\usepackage{booktabs}
\usepackage{url}
\usepackage{subcaption}
\usetikzlibrary{positioning, calc, fit}
\usetikzlibrary{shapes.geometric}
\usetikzlibrary{backgrounds}
\newcommand{\algname}{OATS\;}

\definecolor{pastelBlue}{RGB}{0, 114, 178}    %
\definecolor{pastelYellow}{RGB}{230, 159, 0}  %
\definecolor{pastelRed}{RGB}{204, 51, 51} %
\definecolor{pastelGreen}{RGB}{204, 121, 167} %

\title{OATS: Outlier-Aware Pruning through \\ Sparse and Low Rank Decomposition}

\author{Stephen Zhang \\
University of Toronto\\
\texttt{stephenn.zhang@mail.utoronto.ca}
\And
Vardan Papyan \\
University of Toronto \\
\texttt{vardan.papyan@utoronto.ca}
}

\iclrfinalcopy %
\begin{document}

\maketitle

\begin{abstract}
The recent paradigm shift to large-scale foundation models has brought about a new era for deep learning that, while has found great success in practice, has also been plagued by prohibitively expensive costs in terms of high memory consumption and compute. To mitigate these issues, there has been a concerted effort in post-hoc neural network pruning techniques that do not require costly retraining. Despite the considerable progress being made, existing methods often exhibit a steady drop in model performance as the compression increases. In this paper, we present a novel approach to compressing large transformers, coined OATS, that compresses the model weights by approximating each weight matrix as the sum of a sparse matrix and a low-rank matrix. Prior to the decomposition, the weights are first scaled by the second moment of their input embeddings, so as to ensure the preservation of  outlier features recently observed in large transformer models. Without retraining, OATS achieves state-of-the-art performance when compressing large language models, such as Llama-3 and Phi-3, and vision transformers, such as Google's ViT and DINOv2, by up to $60\%$, all while speeding up the model's inference on a CPU by up to $1.37\times$ compared to prior pruning methods. Our code is available at: \href{https://github.com/stephenqz/OATS}{https://github.com/stephenqz/OATS}. 
\end{abstract}

\section{Introduction}
Large scale transformer-based models have found great success in a number of domains ranging from image classification \citep{wu2020visual}, language modeling \citep{devlin-etal-2019-bert}, and question answering \citep{NEURIPS2020_1457c0d6}. However, these models contain billions of parameters making them computationally expensive to train and deploy, which has lead to an increased demand for resource-saving techniques like model quantization \citep{dettmers2022gptint, egiazarian2024extreme}, parameter efficient fine-tuning \citep{hu2022lora, zhao2024galore}, and, most relevant to this work, neural network pruning \citep{frantar-sparsegpt}. 

Pruning has been a key focus for model compression since the early days of deep neural networks \citep{NIPS1988_07e1cd7d, NIPS1989_6c9882bb, NIPS1992_303ed4c6}. Various pruning techniques have since emerged, introducing sparsity in the model parameters either before training \citep{lee2018snip, Wang2020Picking, NEURIPS2020_46a4378f, jorge2021progressive}, during training \citep{zhugupta, pmlr-v119-evci20a}, or post-training \citep{pmlr-v202-benbaki23a}. In the context of large foundation models, post-training pruning methods, particularly those requiring minimal \citep{xia2022structured, ma2023llmpruner} or no re-training \citep{frantar-sparsegpt, sun2024a, ashkboos2024slicegpt, zhang2024dynamic} are preferred for their computational efficiency. These techniques, when compressing models by $50\%$, have demonstrated the ability to accelerate end-to-end CPU inference by up to $1.8\times$ \citep{yin2024outlier} and GPU inference by up to $1.63\times$ using structured N:M sparsity \citep{Mishra2021AcceleratingSD}, highlighting their potential in reducing costs during deployment.

Despite the significant advancements in pruning techniques, it was recently shown that current methods suffer from a consistent degradation in model performance as compression levels increase \citep{yin2024junk}. Moreover, although structured pruning offers greater potential for acceleration compared to unstructured pruning, it often imposes a much steeper trade-off in terms of model accuracy and effectiveness \citep{pmlr-v162-chen22a}. These challenges underscore the need for more sophisticated pruning strategies that can achieve better performance as compression increases.

\subsection{Contributions}
To mitigate these issues, we introduce \textbf{O}utlier-\textbf{A}ware Pruning \textbf{T}hrough \textbf{S}parse and Low Rank Decomposition (\textbf{OATS}): a novel retraining-free method for compressing large transformers that approximates the model's weight matrices as a sum of a sparse matrix and a low-rank matrix. In order to emphasize the outliers recently observed in large transformer models and preserve model performance \citep{kovaleva-etal-2021-bert, dettmers2022gptint, darcet2024vision, sun2024massive}, OATS first scales the weights by the second moment of their corresponding input embeddings.

We evaluate OATS on recent large language models (LLMs) -- Phi-3 \citep{phi3tr} and Llama-3 \citep{dubey2024llama3herdmodels} -- and vision transformers -- Google's ViT \citep{wu2020visual} and DinoV2 \citep{oquab2023dinov2} -- demonstrating that OATS achieves new state-of-the-art performance across a wide range of commonly employed performance metrics. Furthermore, by combining structured pruning with unstructured pruning, OATS accelerates CPU inference across all levels of compression when compared to models that utilize just unstructured pruning. 

To gain a deeper understanding of the sparse and low-rank terms found by OATS, we split the compressed vision transformers \citep{wu2020visual} into two separate models, a sparse model and a low-rank model, and visualize their respective attention heat maps utilizing attention rollout \citep{abnar-zuidema-2020-quantifying}. These reveal a complementary relationship between the two models, with each focusing on different key areas of the image, effectively segmenting it into distinct regions.

\section{The \algname Algorithm}
The key observation behind the OATS algorithm is that the weight matrices, $\mW\in \mathbb{R}^{d_{out}\times d_{in}}$, in a transformer model can be faithfully approximated as a summation of a sparse and low-rank matrix by solving the following optimization problem, commonly known as Robust PCA \citep{CHANDRASEKARANslm,slr, rpca}:
\begin{equation}\label{eqn:rpca}
    \min_{\mS, \mL \in \mathbb{R}^{d_{out} \times d_{in}}} \| \mW - \mS - \mL \|_F^2 \; \text{ s.t. } \; \text{Rank}(\mL) \leq r, \; \|\mS\|_0 \leq k. 
\end{equation}

\subsection{Alternating Thresholding} To solve Equation \ref{eqn:rpca}, \algname leverages the alternating thresholding algorithms proposed by \citet{godec}, \citet{alt_proj} and \citet{alt_rpca} that iteratively alternates between solving for the low-rank term $\mL$, through singular-value thresholding, and for the sparse term $\mS$, through hard-thresholding. Given a matrix $\mA \in \mathbb{R}^{m\times n}$, singular-value thresholding, also known as truncated SVD, is defined as:
$$ \textsc{TruncatedSVD}(\mA, r) = \mU_r \mathbf{\Sigma}_r \mV_r^{\top},$$
where $\mU_r, \mathbf{\Sigma}_r, \mV_r^{\top}$ correspond to the matrices formed by retaining only the top-$r$ singular vectors and singular values from the full SVD of $\mA$. Hard-thresholding, which succeeds the singular-value thresholding step, is defined as:
$$ \textsc{HardThreshold}(\mA, k) = \mM \odot \mA,$$
\begin{wrapfigure}{r}{0.5\textwidth}
\vspace{-0.5em}
    \begin{minipage}{0.5\textwidth}
    \vspace{-1.5em}
\begin{algorithm}[H]
\caption{\textsc{AlternatingThresholding}} \label{alg:altmin}
\begin{algorithmic}[1]
                \STATE \textbf{Inputs:} 
                \STATE \hspace{1em} Weight Matrix: $\mW \in \mathbb{R}^{d_{out} \times d_{in}}$
                \STATE \hspace{1em} Iterations: $N$
                \STATE \hspace{1em} Rank: $r$
                \STATE \hspace{1em} Nonzeros: $k$
                \vspace{0.5em}
                \STATE \textbf{Procedure:}
                \STATE \hspace{1em} $\mS = \mathbf{0}$
                \STATE \hspace{1em} \textbf{for} $t = 1$ to $N$ \textbf{do}
                    \STATE \hspace{2em} $\mL = \textsc{TruncatedSVD}(\mW - \mS, r)$
                    \STATE \hspace{2em} $\mS = \textsc{HardThreshold}(\mW - \mL, k)$
                \STATE \hspace{1em} \textbf{end for}
                \STATE \hspace{1em} \textbf{return}: $\mS$, $\mL$ 
            \end{algorithmic}
\end{algorithm}
        \vspace{-5.5em}
    \end{minipage}
\end{wrapfigure}
where $\mM \in \mathbb{R}^{m\times n} $ is a binary matrix with $k$ non-zero entries coinciding with the $k$ largest entries in magnitude in $\mA$. These steps are summarized in Algorithm \ref{alg:altmin} on the right. To optimize memory usage, the low-rank term $\mL$ is stored through its two low-rank components: $\mU_r$ and $\mathbf{\Sigma}_r \mV_r^{\top}$. 
\subsection{Alternative Sparsity Patterns}
When performing the hard-threshold step, various restrictions can be enforced on the sparsity pattern of the sparse term of the decomposition for enhanced performance or speed-up. The following are two important cases:
\paragraph{Row-Wise Thresholding} The hard-thresholding can be performed row-wise rather than layer-wise in which case $\mM$ would be a binary matrix with $m\cdot \left \lfloor \frac{k}{m} \right \rfloor$ non-zero entries coinciding with the $\left \lfloor \frac{k}{m} \right \rfloor$ largest entries in magnitude in each row of $\mA$. \citet{sun2024a} have shown this leads to better performance.
\paragraph{N:M Sparsity} The hard-thresholding can be applied at an even more granular level using N:M sparsity, where only the $N$ largest entries by magnitude in every group of $M$ entries in matrix $\mA$ are nonzero. Recently, NVIDIA’s sparse tensor cores have been able to exploit such sparsity patterns for acceleration \citep{Mishra2021AcceleratingSD}.
\subsection{Incorporating Outlier Information} The alternating thresholding on its own yields suboptimal results because the activations of large-scale transformers exhibit a small number of large-magnitude features and altering these (for example, through the sparse and low-rank approximation) negatively impacts model performance \citep{kovaleva-etal-2021-bert, dettmers2022gptint, darcet2024vision, sun2024massive}. OATS takes inspiration from Wanda \citep{sun2024a} and computes a diagonal scaling matrix $\mD\in \mathbb{R}^{d_{in}\times d_{in}}$ that captures the second moment of the input activations
$$ \mD =  \sqrt{\text{diag}(\mX^{\top} \mX)},$$
where $\mX\in \mathbb{R}^{B\times d_{in}}$ and $B$ is the product of the batch size and sequence length. This diagonal matrix, containing large magnitudes for the outlier features, is used to amplify their significance in the reconstruction error of Equation \ref{eqn:rpca}, leading to the following alternative optimization problem:
$$ \min_{\mS, \mL \in \mathbb{R}^{d_{out} \times d_{in}}} \| \mW \mD - \mS - \mL \|_F^2 \; \text{ s.t. } \; \text{Rank}(\mL) \leq r, \; \|\mS\|_0 \leq k. $$
The solution of the problem is given by:
$$ \mS, \mL = \textsc{AlternatingThresholding}\left(\mW \mD, N, r, k\right) $$
which gives a sparse plus low-rank approximation of $\mW \mD \approx \mS+ \mL$. OATS then applies the inverse transformation to reach the final compressed weight:
$$\mW_{\text{compressed}} =  (\mS + \mL)\mD^{-1}, $$
where it leverages the fact that $\mD$ is diagonal so that it both preserves the sparsity pattern of $\mS$ and is easy to invert. The original weight matrix is replaced with three matrices: the sparse matrix $\mS \mD^{-1}$, and two matrices coinciding with the low-rank factorization of $\mL\mD^{-1}$. Aligned with \citet{frantar-sparsegpt, sun2024a}, and \citet{zhang2024dynamic}, the activations are calculated through a calibration set that is propagated through the compressed layers. 

\subsection{OATS Parameters} To determine the rank $r$ and the number of nonzeros $k$, \algname takes in as input two hyperparameters: the \emph{compression rate}, $\rho \in (0,1)$, and the \emph{rank ratio}, $\kappa \in (0,1)$. The compression rate coincides with the sparsity rate required by existing pruning algorithms and is defined as:
$$\rho = 1 - \frac{\# \text{ of nonzero parameters in compressed layer}}{\# \text{ of parameters in original layer}} = 1- \frac{k + r(d_{out} + d_{in})}{d_{out} \cdot d_{in}}. $$ 
The rank ratio represents the proportion of nonzero parameters that appear in the low-rank term: $$\kappa = \frac{\# \text{ of parameters in low-rank term}}{\#\text{ of nonzero parameters in compressed layer}} = \frac{r(d_{out} + d_{in})}{(1-\rho)d_{out} \cdot d_{in}}.$$ 
Given a fixed compression rate $\rho$ and rank ratio $\kappa$, the two equations above can be solved to obtain the rank $r$ and nonzeros $k$: 
\begin{equation}\label{eqn:oats_hyper}
r = \left \lfloor \kappa \cdot (1-\rho) \cdot \frac{d_{out} \cdot d_{in}}{d_{out} + d_{in}} \right \rfloor \qquad k = \left \lfloor (1-\kappa) \cdot (1 - \rho) \cdot d_{out}\cdot d_{in}\right \rfloor.
\end{equation} 
The complete \algname algorithm pseudocode can be found in Algorithm \ref{alg:OATS} below. 
\begin{algorithm}[H]
\caption{\textsc{\algname}} \label{alg:OATS}
\begin{algorithmic}[1]
        \STATE \textbf{Inputs:} 
        \STATE \hspace{1em} Layer Inputs Propagated Through Prior Compressed Layers: $\mX^{\ell} \in \mathbb{R}^{B \times d_{in}}$
        \STATE \hspace{1em} Layer Matrix: $\mW^{\ell} \in \mathbb{R}^{d_{out}\times d_{in}}$
        \STATE \hspace{1em} Compression Rate: $\rho$
        \STATE \hspace{1em} Rank Ratio: $\kappa$
        \STATE \hspace{1em} Iterations: $N$
        \vspace{0.5em}
        \STATE \textbf{Procedure:}
        \STATE \hspace{1em} $r \leftarrow \left \lfloor \kappa \cdot (1-\rho) \cdot \frac{d_{out} \cdot d_{in}}{d_{out} + d_{in}} \right \rfloor, \; k \leftarrow \left \lfloor (1-\kappa) \cdot (1 - \rho) \cdot d_{out}\cdot d_{in}\right \rfloor$
        \STATE \hspace{1em} $\mD \leftarrow \sqrt{\text{diag}(\mX^{\top} \mX)}$ 
        \STATE \hspace{1em} $\mL, \mS \leftarrow \textsc{AlternatingThresholding}(\mW \mD, N, r, k)$
        \STATE \hspace{1em} $\mW \leftarrow (\mL + \mS)\mD^{-1}$ 
        \STATE \hspace{1em} \textbf{return:} $\mX^{\ell+1} \leftarrow \mX^{\ell}\mW^{\top}$
        \end{algorithmic}
    
\end{algorithm}

\section{Experiments on Large Language Models}
\subsection{Experiment Setup}
\paragraph{Models and Tasks} We evaluate \algname on two state-of-the-art families of LLMs: Phi-3 \citep{phi3tr} and Llama-3 \citep{dubey2024llama3herdmodels}. To gauge the algorithm's performance under various model sizes, we select Phi-3 Mini, a $3.8$B parameter model, Phi-3 Medium, a $14$B parameter model, Llama-3 8B, an $8$B parameter model, and Llama-3 70B, a $70$B parameter model. We utilize LM Harness developed by \citet{eval-harness} to evaluate five-shot performance on the Massive Multitask Language Understanding benchmark by \citet{hendrycks2021measuring}, zero-shot performance on eight tasks, and language generation on WikiText-2. 
\begin{wraptable}{r}{7.5cm}
\centering
\begin{tabular}{lccc}
\toprule
\textbf{Parameters} & \textbf{Phi-3} &\textbf{Llama-3} \\ 
\midrule
Iterations  &  80 & 80  \\ 
Rank Ratio  &  25\% & 30\% \\ 
\bottomrule
\end{tabular}
\caption{Hyperparameters utilized for \algname across model families. Both parameters are further ablated in Section \ref{sec:hyper}.} \label{table:oats_hyper}
\vspace{-1.9em}
\end{wraptable}
\vspace{-1.8em}
\paragraph{Pruning Benchmarks} As \algname does not require costly retraining after model compression, we opt to benchmark it with three current state-of-the-art algorithms that similarly do not require such overhead: SparseGPT by \citet{frantar-sparsegpt}, Wanda by \citet{sun2024a}, and DSNoT\footnote{DSNoT experiments are run with both SparseGPT and Wanda. We report the best results across the two. Further details are in Appendix \ref{app:pruning_hyper}.} by \citet{zhang2024dynamic}. The parameters utilized for \algname are depicted in Table \ref{table:oats_hyper}.
\paragraph{Calibration Data} Remaining consistent with \citet{frantar-sparsegpt}, \citet{sun2024a}, and \citet{zhang2024dynamic}, our calibration data consists of 128 sequences of length 2048 sampled from the first shard of the C4 training set \citep{JMLR:v21:20-074}. To ensure consistency, we utilize the same calibration data for all pruning algorithms that we benchmark. 
\paragraph{Layer-Wise Compression Rates} We benchmark our algorithm across a wide range of compression rates: \{0.3, 0.4, 0.5, 0.6\}. For compression rates at or below $0.5$, we compress all transformer blocks uniformly. At the higher compression rate of $0.6$, we utilize Outlier Weighed Layerwise Sparsity Ratios (OWL) proposed by \citet{yin2024outlier} which were shown to lead to significant performance improvements at higher compression rates. All linear layers in a transformer block are pruned uniformly to achieve the desired sparsity rate. We exclude pruning any linear layers that are present in the model head and embeddings which conforms with prior works by \citet{frantar-sparsegpt}, \citet{sun2024a}, and \citet{zhang2024dynamic}. 
\paragraph{Hardware Speedup} We benchmark the CPU speedup of OATS over its competitors using the DeepSparse Inference Engine developed by \citet{deepsparse}. For GPU speed-up, we include structured N:M sparsity experiments where the rank ratio is varied to measure the trade-off between compression and performance. %
\subsection{Results} \label{sec:results_llm}
\paragraph{Five-shot MMLU} Table \ref{table:mmlu}, below, reports the MMLU accuracy of \algname relative to current state-of-the-art pruning algorithms. \algname is able to outperform all prior methods, across all compression rates, with an increasing gap as the compression rate increases. Notably, at $50\%$ compression, \algname surpasses previous pruning algorithms by a margin of $5.42\%$ on Phi-3 Mini, $2.52\%$ on Phi-3 Medium, $2.86\%$ on Llama-3 8B, and $2.03\%$ on Llama-3 70B.
\begin{table}[H]
\centering

\begin{tabular}{lccccc}
\toprule
 \multirow{2}{*}{\textbf{Compression}} & \multirow{2}{*}{\textbf{Method}} & \multicolumn{2}{c}{\textbf{Phi-3}} & \multicolumn{2}{c}{\textbf{Llama-3}}\\ 
\cmidrule(lr){3-6}
 &  & \textbf{Mini ($\mathbf{3.8}$B)} & \textbf{Medium ($\mathbf{14}$B)} & \textbf{8B} & \textbf{70B} \\ 
\midrule
0\% & Dense   & 70.34 & 76.78 & 64.97 & 79.63 \\ 
\midrule 
\multirow{4}{*}{30\%} & SparseGPT  & 68.31 & 74.12 & 64.25 & 78.28  \\ 
& Wanda     &  67.63 & 75.18 & 63.67 & \textbf{79.15}\\ 
& DSNoT     &  68.02 & 75.13 & 63.72 & 79.00 \\ 
& \algname & \textbf{68.84} & \textbf{76.15} & \textbf{65.22} & 78.47 \\
\midrule
\multirow{4}{*}{40\%} & SparseGPT  & 63.47 & 72.42 & 60.91 & 76.29 \\ 
& Wanda     &  64.15 & 73.34 & 60.33 & 77.16\\ 
& DSNoT     &  63.57 & 73.20 & 59.99 & 77.70\\ 
& \algname & \textbf{65.75} & \textbf{74.99} & \textbf{62.46} & \textbf{77.89} \\
\midrule 
\multirow{4}{*}{50\%} & SparseGPT  & 53.22 & 67.63 & 53.60 & 72.47\\ 
& Wanda     &  54.57 & 69.76 & 49.83 & 72.04 \\ 
& DSNoT     &  54.28 & 68.65 & 49.20 & 72.76\\ 
& \algname & \textbf{59.99} & \textbf{72.28} & \textbf{56.46} & \textbf{74.79} \\
\bottomrule
\end{tabular}
\caption{Comparison of average five-shot accuracies (\%) on MMLU under different compression rates.} \label{table:mmlu}
\end{table}
\paragraph{Zero-shot Tasks} Table \ref{table:zero_shot}, below, reports the zero-shot accuracy of OATS relative to current state-of-the-art pruning algorithms averaged across the following eight commonly used tasks: PIQA \citep{piqa}; HellaSwag \citep{zellers2019hellaswag}; Winogrande \citep{winogrande}; OpenBookQA \citep{OpenBookQA2018}; RTE \citep{wang-etal-2018-glue}; BoolQ \citep{clark-etal-2019-boolq}; ARC-e and ARC-c \citep{arcchallenge}. Mirroring the trend observed in the five-shot results, the improvement of \algname over prior pruning algorithms increases with compression, culminating in a $2.05\%$ advantage over prior methods when compressing Phi-3 Mini to $50\%$ of its size.
\begin{table}[H]
\centering

\begin{tabular}{lccccc}
\toprule
 \multirow{2}{*}{\textbf{Compression}} & \multirow{2}{*}{\textbf{Method}} & \multicolumn{2}{c}{\textbf{Phi-3}} & \multicolumn{2}{c}{\textbf{Llama-3}} \\ 
\cmidrule(lr){3-6}
&  & \textbf{Mini ($\mathbf{3.8}$B)} & \textbf{Medium ($\mathbf{14}$B)} & \textbf{8B} & \textbf{70B}\\ 
\midrule
0\% & Dense   & 71.99 & 74.27 & 69.79 & 75.27 \\ 
\midrule 
\multirow{4}{*}{30\%} & SparseGPT  & 70.63 & \textbf{74.53} & 69.08 & 75.07  \\ 
& Wanda     &  70.66 & 74.05 & 68.63 & 75.19 \\ 
& DSNoT     &  71.20 & 74.03 & 68.98 & \textbf{75.54} \\ 
& \algname & \textbf{71.48} & 74.04 & \textbf{69.34} & 75.24 \\
\midrule
\multirow{4}{*}{40\%} & SparseGPT  & 69.18 & 74.40 & 67.58 & 74.63  \\ 
& Wanda     &  68.80 & 73.01 & 67.04 & 74.10 \\ 
& DSNoT     &  69.08 & 72.90 & 66.65 & 74.29\\ 
& \algname & \textbf{70.04} & \textbf{74.46} & \textbf{68.68} & \textbf{74.88}  \\
\midrule 
\multirow{4}{*}{50\%} & SparseGPT  & 66.36 & 73.25 & 64.66 & 73.17  \\ 
& Wanda     &  65.03 & 70.96 & 63.27 & 72.85\\ 
& DSNoT     &  65.33 & 71.12 & 62.74 & 72.91\\ 
& \algname & \textbf{68.41} & \textbf{73.39} & \textbf{65.71} & \textbf{73.30} \\
\bottomrule
\end{tabular}
\caption{Comparison of average zero-shot accuracies (\%) under different compression rates. Task-specific scores can be found in Appendix \ref{app:zero_shot}.} \label{table:zero_shot}
\end{table}

\paragraph{Generation Task} Table \ref{table:ppl}, below, reports the WikiText-2 perplexity of \algname relative to current state-of-the-art pruning algorithms. At $50\%$ compression, \algname results in an $8.49\%$ reduction in perplexity on the larger Phi-3 Medium model, and an even larger $8.99\%$, $9.04\%$, and $9.30\%$ reduction on Phi-3 Mini, Llama-3 8B, and Llama-3 70B respectively. 

\begin{table}[H]
\centering

\begin{tabular}{lccccc}
\toprule
 \multirow{2}{*}{\textbf{Compression}} & \multirow{2}{*}{\textbf{Method}} & \multicolumn{2}{c}{\textbf{Phi-3}} & \multicolumn{2}{c}{\textbf{Llama-3}} \\ 
\cmidrule(lr){3-6}
 &  & \textbf{Mini ($\mathbf{3.8}$B)} & \textbf{Medium ($\mathbf{14}$B)} & \textbf{8B} & \textbf{70B}\\ 
\midrule
0\% & Dense   & 9.50 & 6.21 & 10.17 & 2.68  \\ 
\midrule 
\multirow{4}{*}{30\%} & SparseGPT  & 11.19 & 7.48 & 9.71 & 3.24  \\ 
& Wanda     &  10.71 & 7.28 & 9.39 & 3.28\\ 
& DSNoT     &  10.51 & 7.11 & \textbf{9.36} & 3.27 \\ 
& \algname & \textbf{10.27} & \textbf{6.85} & 9.59 & \textbf{3.07}\\
\midrule
\multirow{4}{*}{40\%} & SparseGPT  & 13.03 & 8.52 & 10.01 & 3.99\\ 
& Wanda     &  12.59 & 8.49 & 9.74 & 4.08\\ 
& DSNoT     &  12.17 & 8.24 & 9.60 & 4.10\\ 
& \algname & \textbf{11.53} & \textbf{7.70} & \textbf{9.24}& \textbf{3.68} \\
\midrule 
\multirow{4}{*}{50\%} & SparseGPT  & 16.80 & 9.89 & 11.95 & 5.27  \\ 
& Wanda     &  17.23 & 10.12 & 12.36 & 5.38 \\ 
& DSNoT     &  16.68 & 9.96 & 12.41 & 5.58 \\ 
& \algname & \textbf{15.18} & \textbf{9.05} & \textbf{10.87} &\textbf{4.78} \\
\bottomrule
\end{tabular}
\caption{Comparison of perplexity (lower is better) on WikiText-2 under different compression rates.} \label{table:ppl}
\end{table}
\begin{wraptable}{r}{7.5cm}
\vspace{-1.9em}
\centering
\begin{tabular}{cccc}
\toprule
 \multirow{2}{*}{\textbf{Method}} & \multicolumn{2}{c}{\textbf{Phi-3}} & \multirow{2}{*}{\textbf{Llama-3 8B}} \\
 \cmidrule(lr){2-3}
 & \textbf{Mini} & \textbf{Medium} &  \\ 
\midrule 
 SparseGPT  & 46.20 & 57.91 & 39.48  \\
 Wanda     &  44.22 & 58.49 & 31.20 \\ 
 DSNoT  & 44.75 & 58.20 & 33.28\\
 \algname & \textbf{52.59} &\textbf{64.10} & \textbf{44.46}  \\
\bottomrule
\end{tabular}
\caption{MMLU accuracy ($\%$) of models compressed by $60\%$ using OWL ratios.} \label{table:hs}
\vspace{-2.5em}
\end{wraptable}
\paragraph{Performance Under High Compression} Table \ref{table:hs}, on the right, is the 5-shot MMLU accuracy of models compressed to a higher compression rate of $60\%$, utilizing OWL ratios \citep{yin2024outlier}. Following the trend above, \algname continues to outperforms prior methods by a margin of $6.39\%$ on Phi-3 Mini, $5.61\%$ on Phi-3 Medium, and $4.98\%$ on Llama-3 8B.

\subsection{Studies and Hyperparameter Exploration}\label{sec:hyper}
We conduct ablation studies for OATS, on Phi-3 Mini  at $40\%$ compression rate with a rank ratio of $20\%$, to quantify the impact of the following design choices:
\begin{itemize}
    \item Scaling the weights by the second moment of the input activations, $\mD$, versus not scaling.
    \item Pruning the weights per each output row in the matrix versus pruning layer-wise. 
\end{itemize}
The results are shown in Table \ref{table:ablate} below:
\begin{table}[H]
\centering
\begin{tabular}{lcccc}
\toprule
 \multicolumn{2}{c}{\textbf{Ablation}} & \textbf{MMLU $(\uparrow)$} & \textbf{Zero-shot $(\uparrow)$} & \textbf{Perplexity $(\downarrow)$} \\ 
\midrule 
\multirow{2}{*}{No Scaling} & Layer-Wise  & 62.46 & 67.58 & 19.21  \\ 
& Row-Wise  & 65.31 & 68.22 & 18.34 \\ 
\midrule
\multirow{2}{*}{Scaling by $\mD$}  & Layer-Wise  & 64.44 & 70.52 & 11.68  \\
& Row-Wise  & \textbf{65.84} & \textbf{70.71} & \textbf{11.50}  \\ 
\bottomrule
\end{tabular}
\caption{Ablation results of OATS on Phi-3-Mini, at $40\%$ compression rate, with a rank ratio of $20\%$. Scaling the weights by the second moment of the input activations and pruning row-wise significantly improves the performance of OATS.} \label{table:ablate}
\end{table}
In addition to the ablations, we perform additional experiments to examine the impact of the rank ratio and the number of iterations on the performance of OATS. Figure \ref{fig:ablation} below shows the results.

\begin{figure*}[h]
    \centering
    \begin{tikzpicture}
\small
\begin{groupplot}[
    group style={
        group size=2 by 1,
        horizontal sep=0.2cm,
    },
    width=0.5\textwidth,
    height=5.0cm,
    ylabel={Accuracy (\%)},
    grid=major,
    domain=0:10,
    samples=100,
    ymin=40, ymax=75, %
    xlabel shift={-1ex},
    ylabel shift={-1ex},
    legend style={
        font=\small,
        at={(1.02,0.5)},
        anchor=west,
        draw=none,
        },
]
\nextgroupplot[xlabel={Rank Ratio},
    xtick={0.1, 0.2, 0.3, 0.4, 0.5, 0.6, 0.7},
    xticklabels={0.1, 0.2, 0.3, 0.4, 0.5, 0.6, 0.7},
    tick style={draw=none},
]
\addplot[mark=*, mark options={fill=pastelBlue!50}, pastelBlue,  mark size=1.8pt]coordinates {
    (0.1, 67.37)
    (0.2, 68.02)
    (0.25, 68.41)
    (0.3, 68.63)
    (0.35, 68.40)
    (0.4, 67.30)
    (0.45, 66.97)
    (0.5, 67.70)
    (0.6, 65.24)
    (0.7, 62.08)
};

\addplot[mark=triangle*, mark options={fill=pastelRed!50}, pastelRed,  mark size=2.5pt] coordinates {
    (0.1, 57.96)
    (0.2, 59.12)
    (0.25, 59.99)
    (0.3, 58.68)
    (0.35, 59.02)
    (0.4, 57.09)
    (0.45, 57.09)
    (0.5, 56.75)
    (0.6, 53.17)
    (0.7, 45.09)
};

\nextgroupplot[xlabel={Number of Iterations}, ylabel={}, yticklabels={},
    tick style={draw=none},
    xtick={1, 20, 40, 60, 80, 100, 120, 140, 160},
    xticklabels={1, 20, 40, 60, 80, 100, 120, 140, 160},
]
\addplot[mark=*, mark options={fill=pastelBlue!50}, pastelBlue,  mark size=1.8pt] coordinates {
    (1, 58.93)
    (5, 66.39)
    (10, 67.59)
    (20, 68.02)
    (40, 68.35)
    (60, 68.27)
    (80, 68.41)
    (100, 67.87)
    (120, 67.48)
    (160, 68.78)
};
\addlegendentry{0-shot}

\addplot[mark=triangle*, mark options={fill=pastelRed!50}, pastelRed,  mark size=2.5pt] coordinates {
    (1, 42.48)
    (5, 53.75)
    (10, 57.18)
    (20, 58.01)
    (40, 59.07)
    (60, 59.99)
    (80, 59.99)
    (100, 59.46)
    (120, 59.51)
    (160, 58.97)
};
\addlegendentry{MMLU}
\end{groupplot}
\end{tikzpicture}
    \caption{\label{fig:ablation}The effect of varying the rank ratio and number of iterations on zero-shot and five-shot accuracy.} 
\end{figure*}
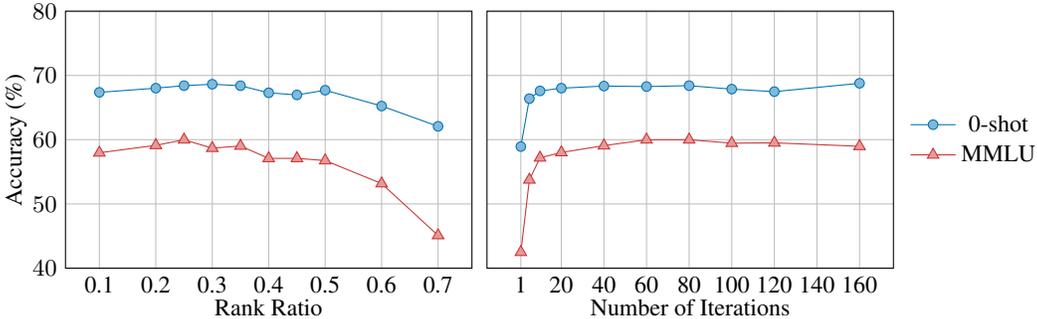

The experiments reveal that a rank ratio between $25\%$ to $30\%$ leads to the best performance, with degradation occurring at higher rank ratios. For the number of iterations, performance improves sharply in the first $20$ iterations, before leveling off and saturating at around $80$ iterations. 
\subsection{Hardware Speedup}\label{sec:cpu_speedup}

\begin{figure*}[h]
    \centering
    \pgfdeclareplotmark{mystarBlue}{
    \node[star,star point ratio=2.25,minimum size=6pt,
          inner sep=0pt,draw=pastelBlue,solid,fill=pastelBlue!50] {};
}
\pgfdeclareplotmark{mystarGreen}{
    \node[star,star point ratio=2.25,minimum size=6pt,
          inner sep=0pt,draw=pastelGreen,solid,fill=pastelGreen!35] {};
}
\pgfdeclareplotmark{mystarYellow}{
    \node[star,star point ratio=2.25,minimum size=6pt,
          inner sep=0pt,draw=pastelYellow,solid,fill=pastelYellow!50] {};
}
\begin{tikzpicture}
\small
\vspace{-3mm}
\begin{groupplot}[
    group style={
        group size=2 by 1,
        horizontal sep=0.2cm,
        vertical sep=0.3cm,
        group name=my plots,
    },
    title style={
        yshift=-1.5ex,
    },
    tick style={draw=none},
    width=0.9\textwidth,
    height=4cm,
    ylabel style={align=center},
    grid=major,
    xmode=log, log basis x=10,
    xtick=data,
    ymin=20,ymax=50,
    xlabel shift={-1ex},
    ylabel shift={-1ex},
    legend style={
        at={(1.02,0.5)}, %
        anchor=west,
        draw=none,
    },
]
\nextgroupplot[
    title={Phi-3 Mini},
    ylabel={MMLU \\Avg. Accuracy},
    xticklabels = {50, 54, 58, 61, 64, 66},
    xlabel={Compression Rate},
    mark size=2.5pt,
    width=0.48\textwidth,
]

\addplot[mark=triangle*, mark options={fill=pastelRed!50}, pastelRed] coordinates {
    (50, 45.82)
    (54.54, 36.96)
    (58.33, 31.11)
    (61.54, 28.16)
    (64.29, 26.12)
    (66.67, 24.04)
};

\addplot[only marks, mark=mystarBlue] coordinates {
    (50, 39.48)
};

\addplot[only marks, mark=mystarYellow] coordinates {
    (50, 32.08)
};

\addplot[only marks, mark=mystarGreen] coordinates {
    (50, 36.96)
};

\nextgroupplot[
    title={Llama-3 8B},
    ylabel={},
    yticklabels={},
    ytick ={30, 40},
    xticklabels = {50, 54, 58, 61, 64, 66},
    xlabel={Compression Rate},
    mark size=2.0pt,
    mark=triangle*,
    width=0.48\textwidth,
]
\addplot[mark=triangle*, mark options={fill=pastelRed!50}, pastelRed] coordinates {
    (50, 43.69)
    (54.54, 37.35)
    (58.33, 30.43)
    (61.54, 28.11)
    (64.29, 25.35)
    (66.67, 25.59)
};

\addplot[only marks, mark=mystarBlue] coordinates {
    (50, 34.49)
};

\addplot[only marks, mark=mystarYellow] coordinates {
    (50, 27.87)
};

\addplot[only marks, mark=mystarGreen] coordinates {
    (50, 29.03)
};

\addlegendentry{OATS}
\addlegendentry{SparseGPT}
\addlegendentry{Wanda}
\addlegendentry{DSNoT}

\end{groupplot}
\end{tikzpicture}
\vspace{-3mm}
    \caption{\label{fig:mn} Experiments evaluating OATS with 2:8 structured sparsity on the sparse terms against 2:4 sparsity of state-of-the pruning algorithms. The rank ratio for OATS is varied to capture the performance across different compression rates. } 
\end{figure*}
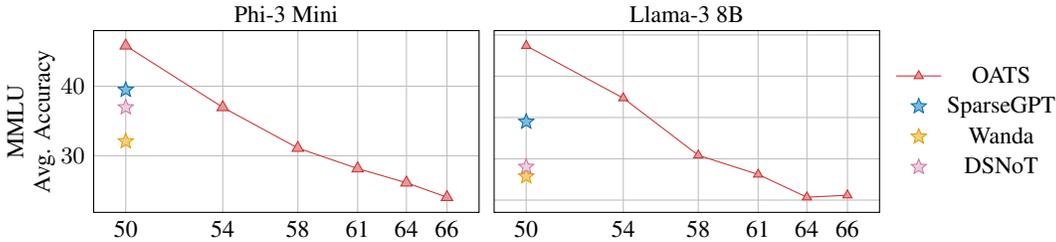

\paragraph{CPU Speedup} We benchmark, using the DeepSparse engine by \citet{deepsparse}, the CPU throughput induced by OATS compared to models pruned with unstructured sparsity. We run end-to-end inference on a compressed Phi-3 Medium 15B model for a single token on an Intel Xeon Gold 6148 CPU @ 2.40GHz with 32 cores (for higher token-counts see Appendix \ref{app:cpu_speedup_2}). The achieved throughput and speedup (over a dense model) are shown in Table \ref{table:cpu_speedup} below. By trading unstructured sparsity for structured sparsity through the low-rank terms, OATS achieves greater CPU speed-up compared to methods that rely solely on unstructured pruning. Notably, at $40\%$ compression, OATS is $1.37\times$ faster than unstructured pruning.  
\begin{table}[H]
\centering
\begin{tabular}{lccc}
\toprule
 \multirow{2}{*}{\textbf{Compression}} & \multirow{2}{*}{\textbf{Method}} & \multirow{2}{*}{\textbf{Throughput}} & \multirow{2}{*}{\textbf{Speedup}} \\ 
 & & &  \\
\midrule 
0\% & Dense &  4.03 & $1.00\times$  \\ 
\midrule
\multirow{2}{*}{30\%} & Unstructured Pruning  &  4.32 & $1.07\times$   \\
& OATS &  \textbf{5.58} &   $\mathbf{1.38\times}$ \\ 
\midrule
\multirow{2}{*}{40\%} & Unstructured Pruning &  5.08 & $1.26\times$   \\
& OATS &  \textbf{6.86} &   $\mathbf{1.73\times}$ \\ 
\midrule
\multirow{2}{*}{50\%} & Unstructured Pruning  &  7.16 & $1.78\times$   \\
& OATS &  \textbf{8.31} &   $\mathbf{2.06\times}$ \\%\\ \midrule
\bottomrule
\end{tabular}
\caption{Comparison of throughput (tokens/second) and speedup achieved through OATS and unstructured pruning methods relative to their dense counterparts.} \label{table:cpu_speedup}
\end{table}
\paragraph{N:M Performance} \label{sec:nm} We compare the performance of state-of-the-art pruning algorithms, using a 2:4 structured sparsity pattern, with the performance of OATS, using a 2:8 structured sparsity pattern on the sparse term. OATS employs a sparser N:M pattern to compensate for its low-rank term that remains dense. We experiment with rank ratios of $\{0.25, 0.3, 0.35, 0.4, 0.45, 0.5\}$. Unlike previous pruning methods, where N:M structured sparsity enforces a fixed compression rate of $\frac{N}{M}$, OATS allows for a flexible trade-off between compression and model performance by adjusting the rank ratio. Figure \ref{fig:mn}, below, illustrates the compression ratio against the 5-shot MMLU accuracy for various compression algorithms.

Despite having a sparser structured sparsity pattern of 2:8, OATS is able to recover the model performance through the presence of its low-rank term. Specifically, at a compression rate of $50\%$, OATS is able to outperform all prior state-of-the-art by $6.34\%$ on Phi-3 Mini. In the case of Llama-3 8B, OATS not only surpasses previous methods by $9.2\%$ at $50\%$ compression, but it also outperforms them by $2.86\%$ at an even higher compression rate of $54\%$.
\section{Experiments on Vision Transformers} \label{sec:vit}
We run experiments on Google's ViT-Base \citep{wu2020visual}, an $86.6$M parameter model trained in a supervised manner on ImageNet-21k \citep{ridnik2021imagenet21k} and fine-tuned on ImageNet 2012 \citep{ILSVRC15}, and DinoV2-Giant \citep{oquab2023dinov2}, a $1.14$B parameter model that was trained through self-supervised learning. 

We benchmark \algname against the same three pruning algorithms: SparseGPT, Wanda, and DSNoT, by evaluating top-1 accuracy on the validation set of ImageNet \citep{ILSVRC15}. A subset of $2048$ images from the training set of ImageNet is used for calibration and is maintained consistent across all pruning experiments. All \algname experiments use a rank ratio of $\kappa {=} 20\%$ and $N{=}80$ iterations. We exclude from compression the embedding and the classifier layers. 

The results are shown in Table 8 below. Compared to LLMs, vision transformers show greater resilience to pruning, with DinoV2 experiencing only a $0.41\%$ drop in top-1 accuracy when compressed by $50\%$ using OATS.
\begin{table}[H]
\centering
\begin{tabular}{lccc}
\toprule
  \textbf{Compression} & \textbf{Method} &\textbf{ViT-Base} & \textbf{DinoV2-Giant} \\
\midrule 
$0\%$ & Dense & 80.33 & 86.55 \\
\midrule
 \multirow{4}{*}{30\%} & SparseGPT  & 80.21 & 86.46 \\
 &Wanda     &  \textbf{80.28} & 86.47 \\ 
 &DSNoT   & 80.16 & 86.46 \\
 &\algname & 80.15 & \textbf{86.52} \\
 \midrule
 \multirow{4}{*}{40\%} & SparseGPT  & 79.58 & 86.39 \\
 &Wanda     &  79.34 & 86.32 \\ 
 &DSNoT   & 79.46 & 86.37 \\
 &\algname & \textbf{79.86} & \textbf{86.46} \\
 \midrule
 \multirow{4}{*}{50\%} & SparseGPT  & 78.44 & 86.04 \\
 &Wanda     &  76.19 & 85.81 \\ 
 &DSNoT   & 76.90 & 85.93 \\
 &\algname & \textbf{78.77} & \textbf{86.14} \\
\bottomrule
\end{tabular}
\caption{ImageNet validation accuracy ($\%$).} \label{table:vit}
\end{table}
\section{Visualizing and Interpreting the Decomposition}\label{sec:rollout}
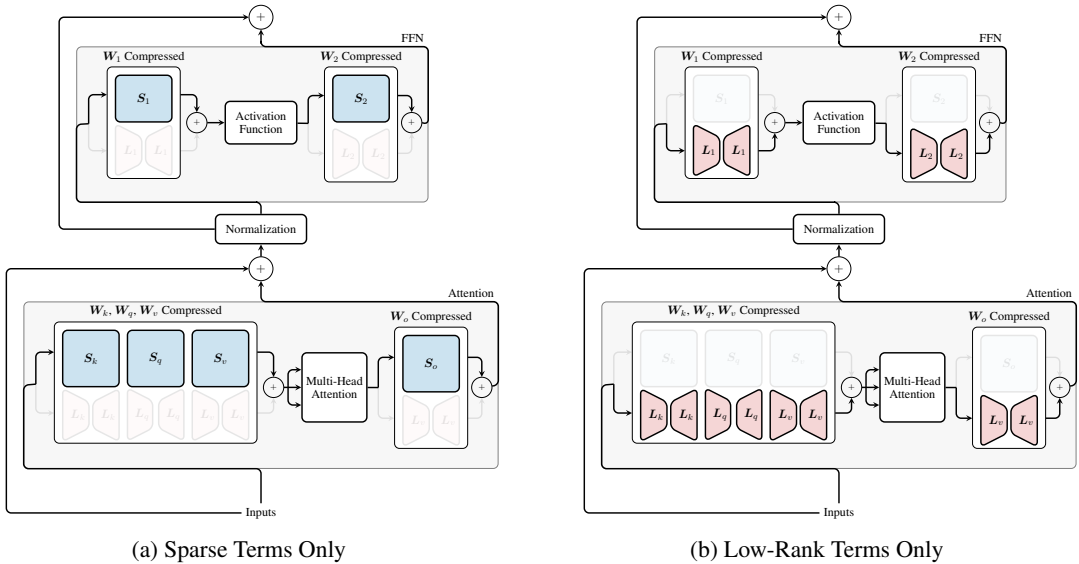
\begin{figure}[h!]
    \centering
    \begin{subfigure}[b]{0.45\textwidth}
        \centering
        \scalebox{0.46}{
            \begin{tikzpicture}[
module/.style={draw, very thick, rounded corners, minimum width=15ex},
linear/.style={module, fill=pastelBlue!20, minimum width=1.63cm, minimum height=1.63cm},
lr_l/.style ={trapezium, trapezium angle=63, shape border rotate=270, minimum width=1.53cm, fill=pastelRed!20, opacity=0.1, draw, very thick, rounded corners},
lr_r/.style ={lr_l, shape border rotate=90},
nonlinearity/.style={module, minimum width=1cm, fill=white},
block/.style={module, fill=black!3, draw=black!50, thin},
comp/.style={module, fill=black!10, thin},
arrow/.style={-stealth, very thick, rounded corners},
fadedarrow/.style={-stealth, opacity=0.1, very thick, rounded corners},
line/.style={very thick, rounded corners},
]
\node[linear](wk_s) {$\mS_k$};
\node[below=4mm of wk_s.south west, anchor=north west, align=left, lr_l, text width=3mm,align=center](wk_l_l) {$\mL_k$};
\node[below=4mm of wk_s.south east, anchor=north east, lr_r, text width=3mm,align=center](wk_l_r) {$\mL_k$};
\node[right=2mm of wk_s, linear](wq_s) {$\mS_q$};
\node[below=4mm of wq_s.south west, anchor=north west, align=left, lr_l, text width=3mm,align=center](wq_l_l) {$\mL_q$};
\node[below=4mm of wq_s.south east, anchor=north east, lr_r, text width=3mm,align=center](wq_l_r) {$\mL_q$};
\node[right=2mm of wq_s, linear](wv_s) {$\mS_v$};
\node[below=4mm of wv_s.south west, anchor=north west, align=left, lr_l, text width=3mm,align=center](wv_l_l) {$\mL_v$};
\node[below=4mm of wv_s.south east, anchor=north east, lr_r, text width=3mm,align=center](wv_l_r) {$\mL_v$};
\node[right=15mm of wv_s.south east, anchor = west, nonlinearity, minimum height=2cm, align=center](mha) {Multi-Head\\Attention};
\node[right=10mm of mha.east, anchor=south west, yshift=-1.5mm, linear](out_s) {$\mS_o$};
\node[below=4mm of out_s.south west, anchor=north west, align=left, lr_l, text width=3mm,align=center](out_l_l) {$\mL_v$};
\node[below=4mm of out_s.south east, anchor=north east, lr_r, text width=3mm,align=center](out_l_r) {$\mL_v$};
\begin{scope}[on background layer]
\node[fit={(wk_s)(wk_l_l)(wk_l_r)(wq_s)(wq_l_l)(wq_l_r)(wv_s)(wv_l_l)(wv_l_r)(mha)(out_s)(out_l_l)(out_l_r)}, block, inner xsep=11mm, inner ysep=8mm] (attnmod) {};
\node[anchor=south east] at (attnmod.north east) {Attention};
\node[fit={(wk_s)(wk_l_l)(wk_l_r)(wq_s)(wq_l_l)(wq_l_r)(wv_s)(wv_l_l)(wv_l_r)}, linear, fill=white, inner sep=2.2mm, thin] (linin) {};
\node[anchor=south] at (linin.north) {$\mW_k$, $\mW_q$, $\mW_v$ Compressed};
\node[fit={(out_s)(out_l_l)(out_l_r)}, linear, fill=white, inner sep=2.2mm, thin] (outpair) {};
\node[anchor=south] at (outpair.north) {$\mW_o$ Compressed};
\end{scope}

\coordinate (lininfork_l) at ($(attnmod.west)!0.4!(linin.west)$);
\draw[fadedarrow] (lininfork_l) |- ($(linin.west)!0.5!(linin.south west)$);
\draw[arrow] (lininfork_l) |- ($(linin.west)!0.5!(linin.north west)$);

\node[draw, circle, left=5mm of mha, minimum size=0.01cm, font=\tiny] (linincomb) {$+$};

\coordinate (mhafork) at ($(mha.west)!0.3!(linin.east)$);
\draw[arrow] (mhafork) |- ($(mha.west)!0.5!(mha.south west)$);
\draw[arrow] (mhafork) |- ($(mha.west)!0.5!(mha.north west)$);
\draw[arrow] (linincomb) -- (mha);
\draw[arrow]  ($(linin.east)!0.5!(linin.north east)$) -| (linincomb);
\draw[fadedarrow]  ($(linin.east)!0.5!(linin.south east)$) -| (linincomb);

\coordinate (outfork_l) at ($(mha.east)!0.4!(outpair.west)$);
\draw[fadedarrow] (outfork_l) |- ($(outpair.west)!0.5!(outpair.south west)$);
\draw[arrow] (outfork_l) |- ($(outpair.west)!0.5!(outpair.north west)$);
\draw[line] (mha) -- (outfork_l);

\node[below=of attnmod] (inputs) {Inputs};
\node[draw, circle, above=0.6cm of attnmod] (resid1) {$+$};
\node[draw, circle, right=3cm of mha, minimum size=0.01cm, font=\tiny] (wocomb) {$+$};

\node[above=0.35cm of resid1, nonlinearity,  align=center, inner sep=3mm](normlayer) {Normalization};

\node[above=2.0cm of normlayer, nonlinearity,  align=center, inner sep=3mm](relu) {Activation\\ Function};
\node[left=15mm of relu.west, anchor=south east, linear, minimum height=1.35cm](w1_s) {$\mS_1$};
\node[draw, circle, left=0.5cm of relu, minimum size=0.01cm, font=\tiny] (w1comb) {$+$};
\node[below=4mm of w1_s.south west, anchor=north west, align=left, lr_l, text width=3mm,align=center](w1_l_l) {$\mL_1$};
\node[below=4mm of w1_s.south east, anchor=north east, lr_r, text width=3mm,align=center](w1_l_r) {$\mL_1$};
\node[right=of relu.east, anchor=south west, linear, minimum height=1.35cm] (w2_s) {$\mS_2$};
\node[below=5mm of w2_s.south west, anchor=north west, lr_l, text width=3mm,align=center](w2_l_l) {$\mL_2$};
\node[below=5mm of w2_s.south east, anchor=north east, lr_r, text width=3mm,align=center](w2_l_r) {$\mL_2$};
\node[draw, circle, right=3cm of relu, minimum size=0.01cm, font=\tiny] (w2comb) {$+$};
\begin{scope}[on background layer]
\node[fit={(w1_s)(w1_l_l)(w1_l_r)(relu)(w2_s)(w2_l_l)(w2_l_r)}, block, inner xsep=11mm, inner ysep=8mm,] (mlp) {};
\node[anchor=south east] at (mlp.north east) {FFN};
\node[fit={(w1_s)(w1_l_l)(w1_l_r)}, linear, fill=white, inner sep=2.2mm, thin] (w1pair){};
\node[anchor=south] at (w1pair.north) {$\mW_1$ Compressed};
\node[fit={(w2_s)(w2_l_l)(w2_l_r)}, linear, fill=white, inner sep=2.2mm, thin] (w2pair) {};
\node[anchor=south] at (w2pair.north) {$\mW_2$ Compressed};
\end{scope}
\coordinate (w1fork_l) at ($(mlp.west)!0.4!(w1pair.west)$);
\draw[fadedarrow] (w1fork_l) |- ($(w1pair.west)!0.5!(w1pair.south west)$);
\draw[arrow] (w1fork_l) |- ($(w1pair.west)!0.5!(w1pair.north west)$);
\draw[arrow]  ($(w1pair.east)!0.5!(w1pair.north east)$) -| (w1comb);
\draw[fadedarrow]  ($(w1pair.east)!0.5!(w1pair.south east)$) -| (w1comb);

\coordinate (w2fork_l) at ($(relu.east)!0.4!(w2pair.west)$);
\draw[fadedarrow] (w2fork_l) |- ($(w2pair.west)!0.5!(w2pair.south west)$);
\draw[arrow] (w2fork_l) |- ($(w2pair.west)!0.5!(w2pair.north west)$);

\draw[arrow] (w1comb) -- (relu);
\draw[line] (relu) -- (w2fork_l);
\coordinate (resid2loc) at ($(resid1.north |- mlp.north) + (0, 0.85cm)$);
\node[draw, circle] (resid2) at (resid2loc) {$+$};

\draw[line] (inputs) -- (attnmod.south) -| (attnmod.west) -- (lininfork_l);
\coordinate[left=5mm of attnmod] (attnleft);
\draw[arrow] (inputs) -| (attnleft) |- (resid1);
\draw[arrow]  ($(outpair.east)!0.5!(outpair.north east)$) -| (wocomb);
\draw[fadedarrow]  ($(outpair.east)!0.5!(outpair.south east)$) -| (wocomb);
\draw[arrow] (wocomb) -- (attnmod.east) |- (attnmod.north) -- (resid1);
\draw[arrow] (resid1) -- (normlayer);
\draw[line] (normlayer.north) |- (mlp.south) -| (mlp.west) -- (w1fork_l);
\coordinate[left=5mm of mlp] (mlpleft);
\draw[arrow] (normlayer) -| (mlpleft) |- (resid2);
\coordinate (mlpleft) at ($(attnleft |- mlp.east)$);
\draw[arrow]  ($(w2pair.east)!0.5!(w2pair.north east)$) -| (w2comb);
\draw[fadedarrow]  ($(w2pair.east)!0.5!(w2pair.south east)$) -| (w2comb);
\draw[arrow] (w2comb.east) -- (mlp.east) |- ($(resid1.north |- mlp.north)$) -| (resid2.south);
\end{tikzpicture}
        }
        \caption{Sparse terms only}
    \end{subfigure}
    \hfill
    \begin{subfigure}[b]{0.45\textwidth}
        \centering
        \scalebox{0.46}{
            \begin{tikzpicture}[
module/.style={draw, very thick, rounded corners, minimum width=15ex},
linear/.style={module, fill=pastelBlue!20, opacity=0.1, minimum width=1.63cm, minimum height=1.63cm},
linearfull/.style={module, fill=pastelBlue!20, minimum width=1.63cm, minimum height=1.63cm},
lr_l/.style ={trapezium, trapezium angle=63, shape border rotate=270, minimum width=1.53cm, fill=pastelRed!20, draw, very thick, rounded corners},
lr_r/.style ={lr_l, shape border rotate=90},
nonlinearity/.style={module, minimum width=1cm, fill=white},
block/.style={module, fill=black!3, draw=black!50, thin},
comp/.style={module, fill=black!10, thin},
arrow/.style={-stealth, very thick, rounded corners},
fadedarrow/.style={-stealth, opacity=0.1, very thick, rounded corners},
line/.style={very thick, rounded corners},
]
\node[linear](wk_s) {$\mS_k$};
\node[below=4mm of wk_s.south west, anchor=north west, align=left, lr_l, text width=3mm,align=center](wk_l_l) {$\mL_k$};
\node[below=4mm of wk_s.south east, anchor=north east, lr_r, text width=3mm,align=center](wk_l_r) {$\mL_k$};
\node[right=2mm of wk_s, linear](wq_s) {$\mS_q$};
\node[below=4mm of wq_s.south west, anchor=north west, align=left, lr_l, text width=3mm,align=center](wq_l_l) {$\mL_q$};
\node[below=4mm of wq_s.south east, anchor=north east, lr_r, text width=3mm,align=center](wq_l_r) {$\mL_q$};
\node[right=2mm of wq_s, linear](wv_s) {$\mS_v$};
\node[below=4mm of wv_s.south west, anchor=north west, align=left, lr_l, text width=3mm,align=center](wv_l_l) {$\mL_v$};
\node[below=4mm of wv_s.south east, anchor=north east, lr_r, text width=3mm,align=center](wv_l_r) {$\mL_v$};
\node[right=15mm of wv_s.south east, anchor = west, nonlinearity, minimum height=2cm, align=center](mha) {Multi-Head\\Attention};
\node[right=10mm of mha.east, anchor=south west, yshift=-1.5mm, linear](out_s) {$\mS_o$};
\node[below=4mm of out_s.south west, anchor=north west, align=left, lr_l, text width=3mm,align=center](out_l_l) {$\mL_v$};
\node[below=4mm of out_s.south east, anchor=north east, lr_r, text width=3mm,align=center](out_l_r) {$\mL_v$};
\begin{scope}[on background layer]
\node[fit={(wk_s)(wk_l_l)(wk_l_r)(wq_s)(wq_l_l)(wq_l_r)(wv_s)(wv_l_l)(wv_l_r)(mha)(out_s)(out_l_l)(out_l_r)}, block, inner xsep=11mm, inner ysep=8mm] (attnmod) {};
\node[anchor=south east] at (attnmod.north east) {Attention};
\node[fit={(wk_s)(wk_l_l)(wk_l_r)(wq_s)(wq_l_l)(wq_l_r)(wv_s)(wv_l_l)(wv_l_r)}, linearfull, fill=white, inner sep=2.2mm, thin] (linin) {};
\node[anchor=south] at (linin.north) {$\mW_k$, $\mW_q$, $\mW_v$ Compressed};
\node[fit={(out_s)(out_l_l)(out_l_r)}, linearfull, fill=white, inner sep=2.2mm, thin] (outpair) {};
\node[anchor=south] at (outpair.north) {$\mW_o$ Compressed};
\end{scope}

\coordinate (lininfork_l) at ($(attnmod.west)!0.4!(linin.west)$);
\draw[arrow] (lininfork_l) |- ($(linin.west)!0.5!(linin.south west)$);
\draw[fadedarrow] (lininfork_l) |- ($(linin.west)!0.5!(linin.north west)$);

\node[draw, circle, left=5mm of mha, minimum size=0.01cm, font=\tiny] (linincomb) {$+$};

\coordinate (mhafork) at ($(mha.west)!0.3!(linin.east)$);
\draw[arrow] (mhafork) |- ($(mha.west)!0.5!(mha.south west)$);
\draw[arrow] (mhafork) |- ($(mha.west)!0.5!(mha.north west)$);
\draw[arrow] (linincomb) -- (mha);
\draw[fadedarrow]  ($(linin.east)!0.5!(linin.north east)$) -| (linincomb);
\draw[arrow]  ($(linin.east)!0.5!(linin.south east)$) -| (linincomb);

\coordinate (outfork_l) at ($(mha.east)!0.4!(outpair.west)$);
\draw[arrow] (outfork_l) |- ($(outpair.west)!0.5!(outpair.south west)$);
\draw[fadedarrow] (outfork_l) |- ($(outpair.west)!0.5!(outpair.north west)$);
\draw[line] (mha) -- (outfork_l);

\node[below=of attnmod] (inputs) {Inputs};
\node[draw, circle, above=0.6cm of attnmod] (resid1) {$+$};
\node[draw, circle, right=3cm of mha, minimum size=0.01cm, font=\tiny] (wocomb) {$+$};

\node[above=0.35cm of resid1, nonlinearity,  align=center, inner sep=3mm](normlayer) {Normalization};

\node[above=2.0cm of normlayer, nonlinearity,  align=center, inner sep=3mm](relu) {Activation\\ Function};
\node[left=15mm of relu.west, anchor=south east, linear, minimum height=1.35cm](w1_s) {$\mS_1$};
\node[draw, circle, left=0.5cm of relu, minimum size=0.01cm, font=\tiny] (w1comb) {$+$};
\node[below=4mm of w1_s.south west, anchor=north west, align=left, lr_l, text width=3mm,align=center](w1_l_l) {$\mL_1$};
\node[below=4mm of w1_s.south east, anchor=north east, lr_r, text width=3mm,align=center](w1_l_r) {$\mL_1$};
\node[right=of relu.east, anchor=south west, linear, minimum height=1.35cm] (w2_s) {$\mS_2$};
\node[below=5mm of w2_s.south west, anchor=north west, lr_l, text width=3mm,align=center](w2_l_l) {$\mL_2$};
\node[below=5mm of w2_s.south east, anchor=north east, lr_r, text width=3mm,align=center](w2_l_r) {$\mL_2$};
\node[draw, circle, right=3cm of relu, minimum size=0.01cm, font=\tiny] (w2comb) {$+$};
\begin{scope}[on background layer]
\node[fit={(w1_s)(w1_l_l)(w1_l_r)(relu)(w2_s)(w2_l_l)(w2_l_r)}, block, inner xsep=11mm, inner ysep=8mm,] (mlp) {};
\node[anchor=south east] at (mlp.north east) {FFN};
\node[fit={(w1_s)(w1_l_l)(w1_l_r)}, linearfull, fill=white, inner sep=2.2mm, thin] (w1pair){};
\node[anchor=south] at (w1pair.north) {$\mW_1$ Compressed};
\node[fit={(w2_s)(w2_l_l)(w2_l_r)}, linearfull, fill=white, inner sep=2.2mm, thin] (w2pair) {};
\node[anchor=south] at (w2pair.north) {$\mW_2$ Compressed};
\end{scope}
\coordinate (w1fork_l) at ($(mlp.west)!0.4!(w1pair.west)$);
\draw[arrow] (w1fork_l) |- ($(w1pair.west)!0.5!(w1pair.south west)$);
\draw[fadedarrow] (w1fork_l) |- ($(w1pair.west)!0.5!(w1pair.north west)$);
\draw[fadedarrow]  ($(w1pair.east)!0.5!(w1pair.north east)$) -| (w1comb);
\draw[arrow]  ($(w1pair.east)!0.5!(w1pair.south east)$) -| (w1comb);

\coordinate (w2fork_l) at ($(relu.east)!0.4!(w2pair.west)$);
\draw[arrow] (w2fork_l) |- ($(w2pair.west)!0.5!(w2pair.south west)$);
\draw[fadedarrow] (w2fork_l) |- ($(w2pair.west)!0.5!(w2pair.north west)$);

\draw[arrow] (w1comb) -- (relu);
\draw[line] (relu) -- (w2fork_l);
\coordinate (resid2loc) at ($(resid1.north |- mlp.north) + (0, 0.85cm)$);
\node[draw, circle] (resid2) at (resid2loc) {$+$};

\draw[line] (inputs) -- (attnmod.south) -| (attnmod.west) -- (lininfork_l);
\coordinate[left=5mm of attnmod] (attnleft);
\draw[arrow] (inputs) -| (attnleft) |- (resid1);
\draw[fadedarrow]  ($(outpair.east)!0.5!(outpair.north east)$) -| (wocomb);
\draw[arrow]  ($(outpair.east)!0.5!(outpair.south east)$) -| (wocomb);
\draw[arrow] (wocomb) -- (attnmod.east) |- (attnmod.north) -- (resid1);
\draw[arrow] (resid1) -- (normlayer);
\draw[line] (normlayer.north) |- (mlp.south) -| (mlp.west) -- (w1fork_l);
\coordinate[left=5mm of mlp] (mlpleft);
\draw[arrow] (normlayer) -| (mlpleft) |- (resid2);
\coordinate (mlpleft) at ($(attnleft |- mlp.east)$);
\draw[fadedarrow]  ($(w2pair.east)!0.5!(w2pair.north east)$) -| (w2comb);
\draw[arrow]  ($(w2pair.east)!0.5!(w2pair.south east)$) -| (w2comb);
\draw[arrow] (w2comb.east) -- (mlp.east) |- ($(resid1.north |- mlp.north)$) -| (resid2.south);
\end{tikzpicture}  %
        }
        \caption{Low-rank terms only}
    \end{subfigure}
    \caption{A visualization of how the attention rollout is computed to isolate the contribution of the sparse terms versus low-rank terms given by the OATS algorithm.}
    \label{fig:low_sparse_viz}
\end{figure}

To develop a better understanding of how the sparse and low rank components individually contribute to the flow of information through the model, we compute and visualize the attention rollout \citep{abnar-zuidema-2020-quantifying} of the compressed vision transformers when:
\begin{itemize}
    \item All low-rank terms are set to zero and inputs are propagated through only the sparse terms.
    \item All sparse terms are set to zero and inputs are propagated through only the low-rank terms.
\end{itemize}
Figure \ref{fig:low_sparse_viz} below provides a visualization of how the information would flow through a standard transformer block for both settings.  

\begin{figure}[h!]
    \begin{tabular}{m{1.6cm} m{1.7cm} m{1.7cm} m{1.7cm} m{1.7cm} m{1.7cm} m{1.7cm}}
    Original: & 
        \includegraphics[width=0.12\textwidth]{vit_viz/og_0.png} & 
        \includegraphics[width=0.12\textwidth]{vit_viz/og_1.png} & 
        \includegraphics[width=0.12\textwidth]{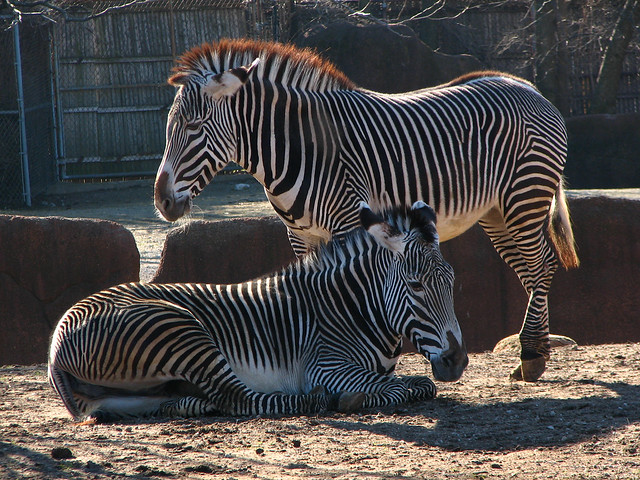} & 
        \includegraphics[width=0.12\textwidth]{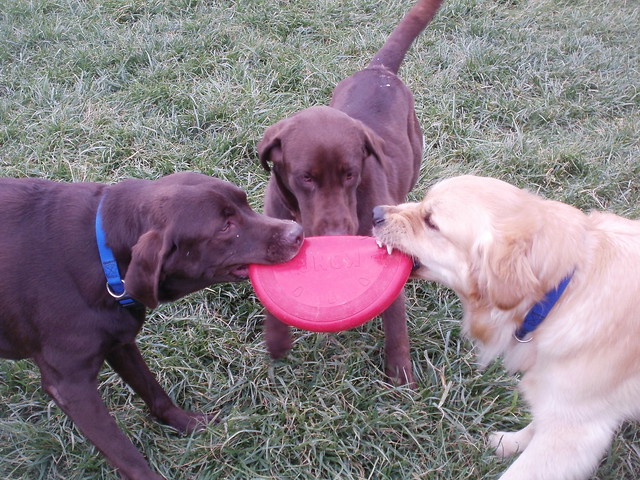} & 
        \includegraphics[width=0.12\textwidth]{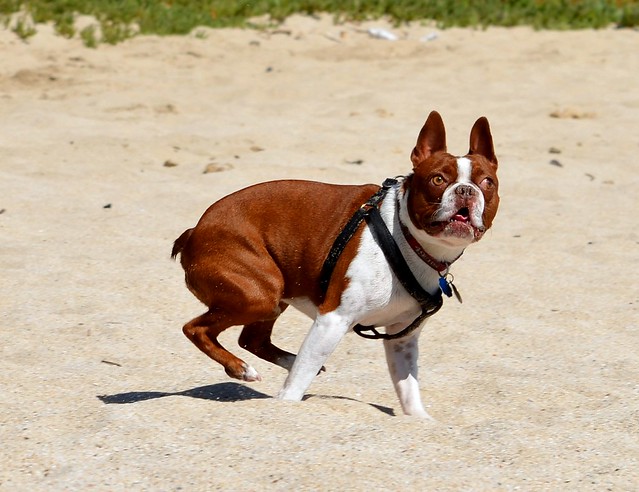} & 
        \includegraphics[width=0.12\textwidth]{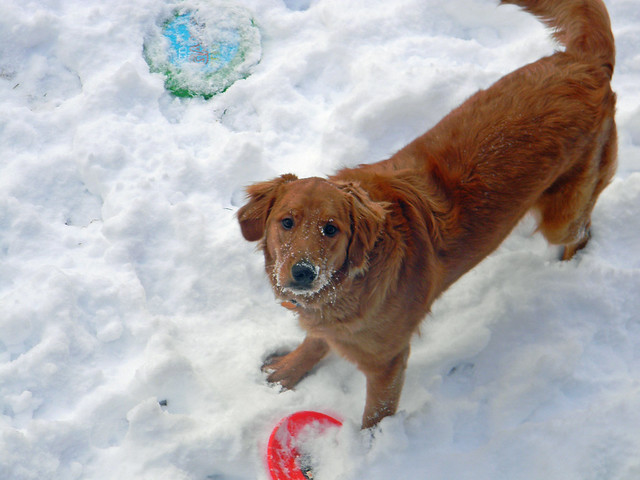} \\
        
        Sparse: & 
        \includegraphics[width=0.12\textwidth]{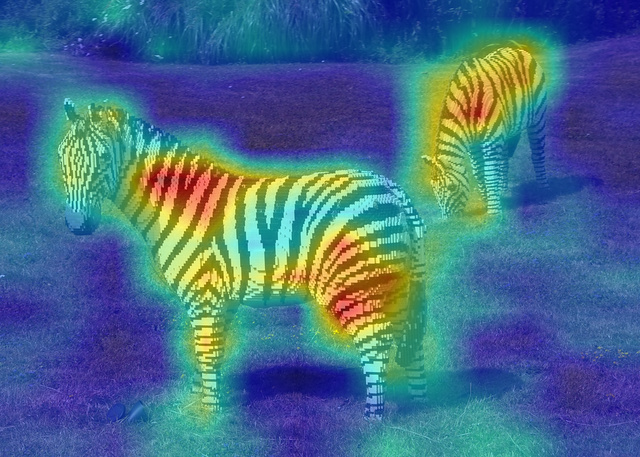} & 
        \includegraphics[width=0.12\textwidth]{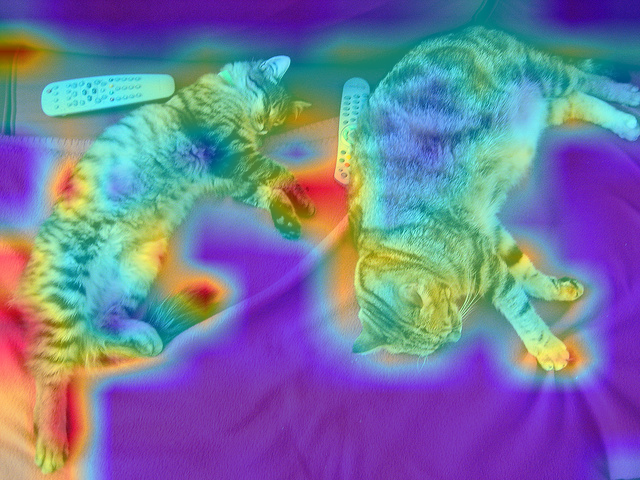} & 
        \includegraphics[width=0.12\textwidth]{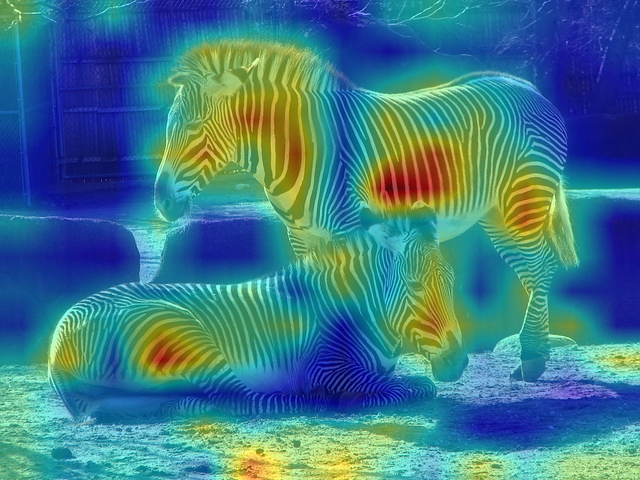} & 
        \includegraphics[width=0.12\textwidth]{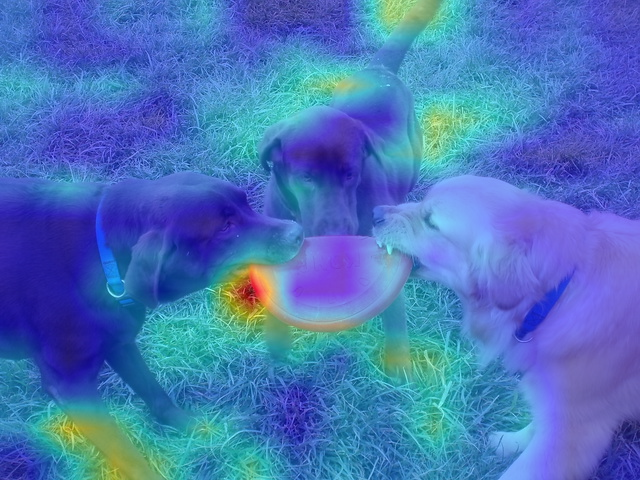} & 
        \includegraphics[width=0.12\textwidth]{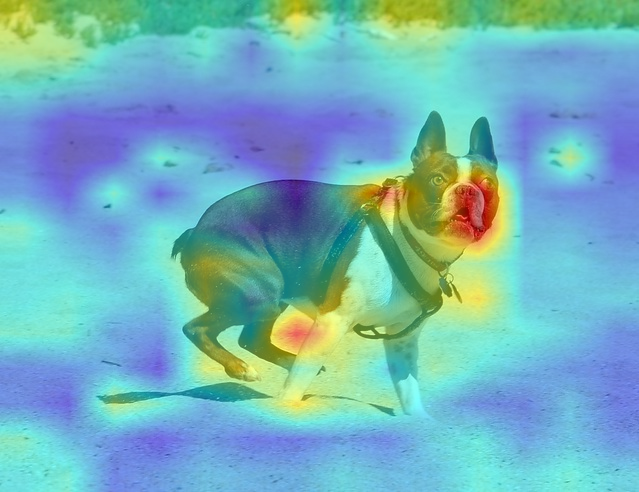} & 
        \includegraphics[width=0.12\textwidth]{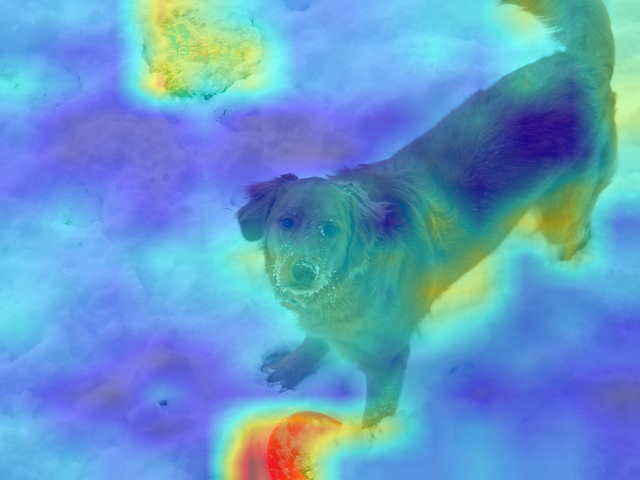} \\
        
        Low-Rank: & 
        \includegraphics[width=0.12\textwidth]{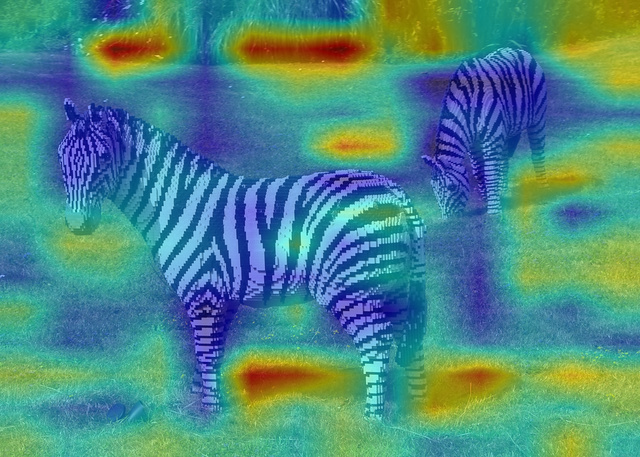} & 
        \includegraphics[width=0.12\textwidth]{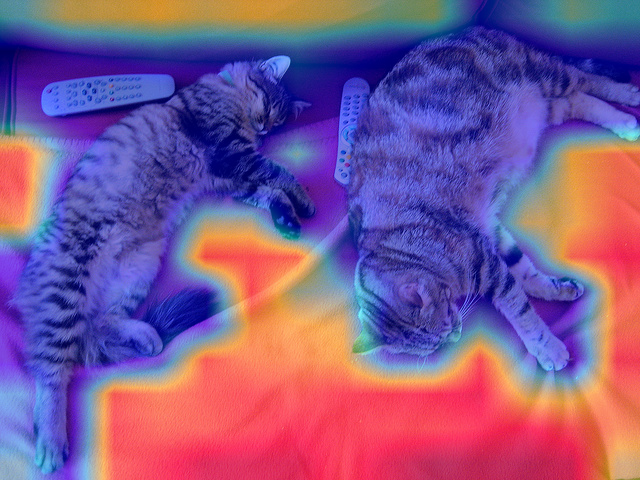} & 
        \includegraphics[width=0.12\textwidth]{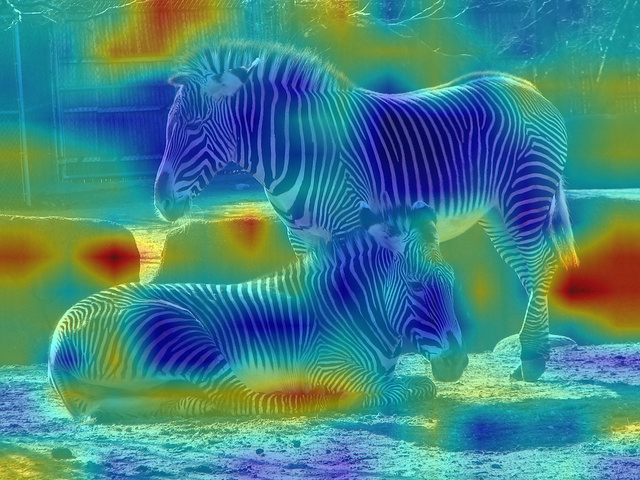} & 
        \includegraphics[width=0.12\textwidth]{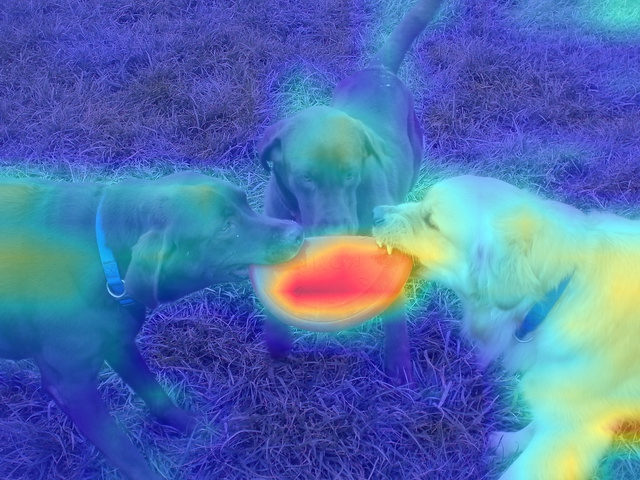} & 
        \includegraphics[width=0.12\textwidth]{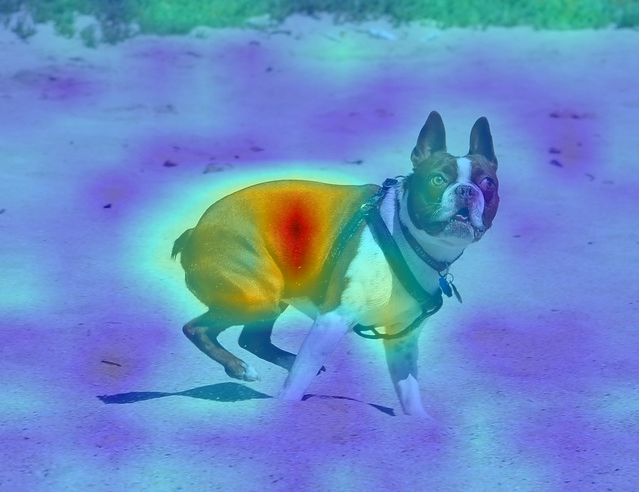} & 
        \includegraphics[width=0.12\textwidth]{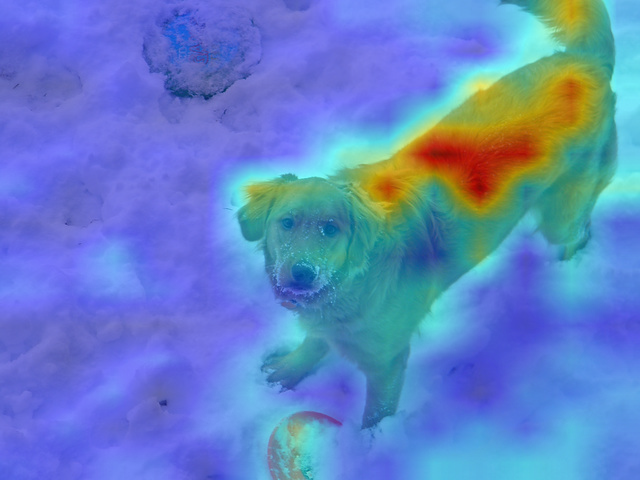} \\
    \end{tabular}
    
    \caption{Attention rollout visualization applied to various images on the Microsoft COCO dataset. }
    \label{fig:rollout_viz}
\end{figure}

Figure \ref{fig:rollout_viz} depicts the attention rollout for various images in the Microsoft COCO dataset \citep{coco} passed to a ViT-B that was compressed by $50\%$, with a rank ratio of $20\%$.

The rollout visualizations show that the sparse and low-rank terms capture distinct areas of the image, effectively segmenting it. A careful analysis reveals three distinct partitioning patterns. The first, which is also commonly exhibited in the classical setting \citep{rpca}, is when one component (commonly sparse) captures the subject(s), while the other component (commonly low-rank) captures the background. The second is when both components focus on different parts of the same subject, each capturing distinct features. The third behavior arises when the image contains multiple subjects, with each component isolating a different subject. While these patterns provide initial insights into how the components process visual information, further investigation is needed to fully understand the mechanisms driving these behaviors.
\section{Related Works}
\paragraph{Connection with Wanda} OATS utilizes the same outlier scaling as the the one that is employed by Wanda \citep{sun2024a}. In fact, Wanda can be seen as a special case of OATS when the rank ratio $\kappa{=}0$. Indeed, in such a case, according to Equation \ref{eqn:oats_hyper}, the low-rank term would become the zero matrix and OATS would perform a single hard thresholding step that is equivalent to the pruning step described by Wanda: $\mW_{\text{compressed}} = \textsc{HardThreshold}(\mW \mD, k) \mD^{-1}$. %
\paragraph{Sparse and Low-Rank Approximation in Transformers} The emergence of sparse and low-rank structures in transformers has recently become an area of both theoretical and practical interest. On the theoretical front, \citet{zhao2024implicit} showed that the logits of LLMs trained utilizing next token prediction converge to a low rank and sparse structure. On the practical front, Scatterbrain proposed by \citet{scatterbrain} shows that it is possible to approximate the entire attention mechanism with a single sparse and low rank decomposition. Pruning-wise, LoRAP by \citet{li2024lorap} performs \emph{structured} pruning on the feed-forward linear layers and apply a low-rank decomposition to the attention matrices using a scaling technique similar to OATS.

\paragraph{Structured Pruning and Low-Rank Adaptation} Recent works, such as LoSparse \citep{losparse}, LoRAPrune \citep{zhang2024loraprune}, and APT \citep{zhao2024apt}, propose variations of applying structured pruning on the weights while incorporating a low-rank adapter that is trained via gradient descent. These are markedly different than OATS, which does not employ any fine-tuning with low-rank adapters, nor does it perform structured pruning (but rather a sparse plus low-rank decomposition which can be thought of as a combination of structured and unstructured pruning).
\paragraph{Robust PCA Algorithms} The search for Robust PCA algorithms has been a key area of interest since the inception of the problem. Examples of other approaches include applying a convex relaxation, where the sparsity and low-rank constraints are replaced by $\ell_1$ and nuclear norm surrogates \citep{spcp}, or parameterizing the low-rank matrix as $\mL = \mU\mV^{\top}$, and applying gradient descent on $\mU$ and $\mV$ \citep{fast_rpca, scale_rpca}. While OATS utilizes the alternating thresholding approach for its simplicity, future work might want to investigate the use of other algorithms.
\paragraph{Pruning and Interpretability}  An active area of research is understanding what pruning is pruning and how it impacts model performance. \citet{Paganini2020PruneR} show that pruning has a disproportionate negative effect on underrepresented classes. In a similar vein, \citet{yin2024junk} showed that pruning LLMs can irreversibly harm model performance on tasks that are more challenging. We postulate that the low-rank term present in OATS might be able to mitigate the negative impacts of pruning. Indeed, Tables \ref{table:mmlu} and \ref{table:hs} show that the gap between OATS and prior methods is larger at higher compression, suggesting that the low-rank term plays a critical role in mitigating the loss in performance. 
\section{Conclusion}
We have introduced OATS, an algorithm that without any re-training, compresses the model's weight matrices through a sparse and low-rank decomposition. Taking inspiration from prior works on the emergence of outlier features, OATS first scales the weights by the second moment of their input embeddings prior to applying an alternating thresholding algorithm. A comprehensive evaluation shows that OATS is able to consistently outperform prior state-of-the-art on various performance metrics across multiple compression rates, models, and modalities, while also improving on CPU speed-up. Beyond just model compression, our visualizations on vision transformers indicate that models exhibit sparse and low-rank structures that capture different segments of the image. This work is the first to reveal the potential of sparse and low-rank decompositions for large-scale transformers, setting the stage for future innovations that can harness this structure to improve model efficiency, performance, and interpretability.

\subsubsection*{Acknowledgments}
We acknowledge the support of the Natural Sciences and
Engineering Research Council of Canada (NSERC). This research was supported in part by the Province of Ontario, the Government of Canada through CIFAR, and industry sponsors of the Vector Institute (\url{www.vectorinstitute.ai/partnerships/current-partners/}). This research was also enabled in part by support provided by Compute Ontario (\url{https://www.computeontario.ca}) and the Digital Research Alliance of Canada (\url{https://alliancecan.ca}).

\bibliography{main}

\begin{thebibliography}{77}
\providecommand{\natexlab}[1]{#1}
\providecommand{\url}[1]{\texttt{#1}}
\expandafter\ifx\csname urlstyle\endcsname\relax
  \providecommand{\doi}[1]{doi: #1}\else
  \providecommand{\doi}{doi: \begingroup \urlstyle{rm}\Url}\fi

\bibitem[Abdin et~al.(2024)Abdin, Jacobs, Awan, Aneja, Awadallah, Awadalla, Bach, Bahree, Bakhtiari, Bao, Behl, Benhaim, Bilenko, Bjorck, Bubeck, Cai, Cai, Mendes, Chen, Chaudhary, Chen, Chen, Chen, Chen, Chopra, Dai, Giorno, de~Rosa, Dixon, Eldan, Fragoso, Iter, Gao, Gao, Gao, Garg, Goswami, Gunasekar, Haider, Hao, Hewett, Huynh, Javaheripi, Jin, Kauffmann, Karampatziakis, Kim, Khademi, Kurilenko, Lee, Lee, Li, Li, Liang, Liden, Liu, Liu, Liu, Lin, Lin, Luo, Madan, Mazzola, Mitra, Modi, Nguyen, Norick, Patra, Perez-Becker, Portet, Pryzant, Qin, Radmilac, Rosset, Roy, Ruwase, Saarikivi, Saied, Salim, Santacroce, Shah, Shang, Sharma, Shukla, Song, Tanaka, Tupini, Wang, Wang, Wang, Wang, Ward, Wang, Witte, Wu, Wyatt, Xiao, Xu, Xu, Xu, Yadav, Yang, Yang, Yang, Yang, Yu, Yuan, Zhang, Zhang, Zhang, Zhang, Zhang, Zhang, Zhang, and Zhou]{phi3tr}
Marah Abdin, Sam~Ade Jacobs, Ammar~Ahmad Awan, Jyoti Aneja, Ahmed Awadallah, Hany Awadalla, Nguyen Bach, Amit Bahree, Arash Bakhtiari, Jianmin Bao, Harkirat Behl, Alon Benhaim, Misha Bilenko, Johan Bjorck, Sébastien Bubeck, Qin Cai, Martin Cai, Caio César~Teodoro Mendes, Weizhu Chen, Vishrav Chaudhary, Dong Chen, Dongdong Chen, Yen-Chun Chen, Yi-Ling Chen, Parul Chopra, Xiyang Dai, Allie~Del Giorno, Gustavo de~Rosa, Matthew Dixon, Ronen Eldan, Victor Fragoso, Dan Iter, Mei Gao, Min Gao, Jianfeng Gao, Amit Garg, Abhishek Goswami, Suriya Gunasekar, Emman Haider, Junheng Hao, Russell~J. Hewett, Jamie Huynh, Mojan Javaheripi, Xin Jin, Piero Kauffmann, Nikos Karampatziakis, Dongwoo Kim, Mahoud Khademi, Lev Kurilenko, James~R. Lee, Yin~Tat Lee, Yuanzhi Li, Yunsheng Li, Chen Liang, Lars Liden, Ce~Liu, Mengchen Liu, Weishung Liu, Eric Lin, Zeqi Lin, Chong Luo, Piyush Madan, Matt Mazzola, Arindam Mitra, Hardik Modi, Anh Nguyen, Brandon Norick, Barun Patra, Daniel Perez-Becker, Thomas Portet, Reid Pryzant, Heyang Qin, Marko Radmilac, Corby Rosset, Sambudha Roy, Olatunji Ruwase, Olli Saarikivi, Amin Saied, Adil Salim, Michael Santacroce, Shital Shah, Ning Shang, Hiteshi Sharma, Swadheen Shukla, Xia Song, Masahiro Tanaka, Andrea Tupini, Xin Wang, Lijuan Wang, Chunyu Wang, Yu~Wang, Rachel Ward, Guanhua Wang, Philipp Witte, Haiping Wu, Michael Wyatt, Bin Xiao, Can Xu, Jiahang Xu, Weijian Xu, Sonali Yadav, Fan Yang, Jianwei Yang, Ziyi Yang, Yifan Yang, Donghan Yu, Lu~Yuan, Chengruidong Zhang, Cyril Zhang, Jianwen Zhang, Li~Lyna Zhang, Yi~Zhang, Yue Zhang, Yunan Zhang, and Xiren Zhou.
\newblock Phi-3 technical report: A highly capable language model locally on your phone, 2024.
\newblock URL \url{https://arxiv.org/abs/2404.14219}.

\bibitem[Abnar \& Zuidema(2020)Abnar and Zuidema]{abnar-zuidema-2020-quantifying}
Samira Abnar and Willem Zuidema.
\newblock Quantifying attention flow in transformers.
\newblock In Dan Jurafsky, Joyce Chai, Natalie Schluter, and Joel Tetreault (eds.), \emph{Proceedings of the 58th Annual Meeting of the Association for Computational Linguistics}, pp.\  4190--4197, Online, July 2020. Association for Computational Linguistics.
\newblock \doi{10.18653/v1/2020.acl-main.385}.
\newblock URL \url{https://aclanthology.org/2020.acl-main.385}.

\bibitem[Ashkboos et~al.(2024)Ashkboos, Croci, do~Nascimento, Hoefler, and Hensman]{ashkboos2024slicegpt}
Saleh Ashkboos, Maximilian~L. Croci, Marcelo~Gennari do~Nascimento, Torsten Hoefler, and James Hensman.
\newblock Slice{GPT}: Compress large language models by deleting rows and columns.
\newblock In \emph{The Twelfth International Conference on Learning Representations}, 2024.
\newblock URL \url{https://openreview.net/forum?id=vXxardq6db}.

\bibitem[Benbaki et~al.(2023)Benbaki, Chen, Meng, Hazimeh, Ponomareva, Zhao, and Mazumder]{pmlr-v202-benbaki23a}
Riade Benbaki, Wenyu Chen, Xiang Meng, Hussein Hazimeh, Natalia Ponomareva, Zhe Zhao, and Rahul Mazumder.
\newblock Fast as {CHITA}: Neural network pruning with combinatorial optimization.
\newblock In \emph{Proceedings of the 40th International Conference on Machine Learning}, pp.\  2031--2049, 2023.

\bibitem[Bertsimas et~al.(2024)Bertsimas, Cory-Wright, and Johnson]{alt_rpca}
Dimitris Bertsimas, Ryan Cory-Wright, and Nicholas A.~G. Johnson.
\newblock Sparse plus low rank matrix decomposition: a discrete optimization approach.
\newblock \emph{J. Mach. Learn. Res.}, 24\penalty0 (1), mar 2024.
\newblock ISSN 1532-4435.

\bibitem[Bisk et~al.(2020)Bisk, Zellers, Le~bras, Gao, and Choi]{piqa}
Yonatan Bisk, Rowan Zellers, Ronan Le~bras, Jianfeng Gao, and Yejin Choi.
\newblock Piqa: Reasoning about physical commonsense in natural language.
\newblock \emph{Proceedings of the AAAI Conference on Artificial Intelligence}, 34\penalty0 (05):\penalty0 7432--7439, Apr. 2020.
\newblock \doi{10.1609/aaai.v34i05.6239}.
\newblock URL \url{https://ojs.aaai.org/index.php/AAAI/article/view/6239}.

\bibitem[Brown et~al.(2020)Brown, Mann, Ryder, Subbiah, Kaplan, Dhariwal, Neelakantan, Shyam, Sastry, Askell, Agarwal, Herbert-Voss, Krueger, Henighan, Child, Ramesh, Ziegler, Wu, Winter, Hesse, Chen, Sigler, Litwin, Gray, Chess, Clark, Berner, McCandlish, Radford, Sutskever, and Amodei]{NEURIPS2020_1457c0d6}
Tom Brown, Benjamin Mann, Nick Ryder, Melanie Subbiah, Jared~D Kaplan, Prafulla Dhariwal, Arvind Neelakantan, Pranav Shyam, Girish Sastry, Amanda Askell, Sandhini Agarwal, Ariel Herbert-Voss, Gretchen Krueger, Tom Henighan, Rewon Child, Aditya Ramesh, Daniel Ziegler, Jeffrey Wu, Clemens Winter, Chris Hesse, Mark Chen, Eric Sigler, Mateusz Litwin, Scott Gray, Benjamin Chess, Jack Clark, Christopher Berner, Sam McCandlish, Alec Radford, Ilya Sutskever, and Dario Amodei.
\newblock Language models are few-shot learners.
\newblock In H.~Larochelle, M.~Ranzato, R.~Hadsell, M.F. Balcan, and H.~Lin (eds.), \emph{Advances in Neural Information Processing Systems}, volume~33, pp.\  1877--1901. Curran Associates, Inc., 2020.
\newblock URL \url{https://proceedings.neurips.cc/paper_files/paper/2020/file/1457c0d6bfcb4967418bfb8ac142f64a-Paper.pdf}.

\bibitem[Cand\`{e}s et~al.(2011)Cand\`{e}s, Li, Ma, and Wright]{rpca}
Emmanuel~J. Cand\`{e}s, Xiaodong Li, Yi~Ma, and John Wright.
\newblock Robust principal component analysis?
\newblock \emph{J. ACM}, 58\penalty0 (3), jun 2011.
\newblock ISSN 0004-5411.
\newblock \doi{10.1145/1970392.1970395}.
\newblock URL \url{https://doi.org/10.1145/1970392.1970395}.

\bibitem[Chandrasekaran et~al.(2009)Chandrasekaran, Sanghavi, Parrilo, and Willsky]{CHANDRASEKARANslm}
Venkat Chandrasekaran, Sujay Sanghavi, Pablo~A. Parrilo, and Alan~S. Willsky.
\newblock Sparse and low-rank matrix decompositions.
\newblock \emph{IFAC Proceedings Volumes}, 42\penalty0 (10):\penalty0 1493--1498, 2009.
\newblock ISSN 1474-6670.
\newblock \doi{https://doi.org/10.3182/20090706-3-FR-2004.00249}.
\newblock URL \url{https://www.sciencedirect.com/science/article/pii/S1474667016388632}.
\newblock 15th IFAC Symposium on System Identification.

\bibitem[Chandrasekaran et~al.(2011)Chandrasekaran, Sanghavi, Parrilo, and Willsky]{slr}
Venkat Chandrasekaran, Sujay Sanghavi, Pablo~A. Parrilo, and Alan~S. Willsky.
\newblock Rank-sparsity incoherence for matrix decomposition.
\newblock \emph{SIAM Journal on Optimization}, 21\penalty0 (2):\penalty0 572--596, 2011.
\newblock \doi{10.1137/090761793}.
\newblock URL \url{https://doi.org/10.1137/090761793}.

\bibitem[Chavan et~al.(2022)Chavan, Shen, Liu, Liu, Cheng, and Xing]{chavan2022vision}
Arnav Chavan, Zhiqiang Shen, Zhuang Liu, Zechun Liu, Kwang-Ting Cheng, and Eric Xing.
\newblock Vision transformer slimming: Multi-dimension searching in continuous optimization space.
\newblock 2022.

\bibitem[Chen et~al.(2021{\natexlab{a}})Chen, Dao, Winsor, Song, Rudra, and R\'{e}]{scatterbrain}
Beidi Chen, Tri Dao, Eric Winsor, Zhao Song, Atri Rudra, and Christopher R\'{e}.
\newblock Scatterbrain: Unifying sparse and low-rank attention.
\newblock In M.~Ranzato, A.~Beygelzimer, Y.~Dauphin, P.S. Liang, and J.~Wortman Vaughan (eds.), \emph{Advances in Neural Information Processing Systems}, volume~34, pp.\  17413--17426. Curran Associates, Inc., 2021{\natexlab{a}}.
\newblock URL \url{https://proceedings.neurips.cc/paper_files/paper/2021/file/9185f3ec501c674c7c788464a36e7fb3-Paper.pdf}.

\bibitem[Chen et~al.(2021{\natexlab{b}})Chen, Cheng, Gan, Yuan, Zhang, and Wang]{chen2021chasing}
Tianlong Chen, Yu~Cheng, Zhe Gan, Lu~Yuan, Lei Zhang, and Zhangyang Wang.
\newblock Chasing sparsity in vision transformers: An end-to-end exploration.
\newblock In A.~Beygelzimer, Y.~Dauphin, P.~Liang, and J.~Wortman Vaughan (eds.), \emph{Advances in Neural Information Processing Systems}, 2021{\natexlab{b}}.
\newblock URL \url{https://openreview.net/forum?id=LKoMTwTuQnC}.

\bibitem[Chen et~al.(2022)Chen, Chen, Ma, Wang, and Wang]{pmlr-v162-chen22a}
Tianlong Chen, Xuxi Chen, Xiaolong Ma, Yanzhi Wang, and Zhangyang Wang.
\newblock Coarsening the granularity: Towards structurally sparse lottery tickets.
\newblock In Kamalika Chaudhuri, Stefanie Jegelka, Le~Song, Csaba Szepesvari, Gang Niu, and Sivan Sabato (eds.), \emph{Proceedings of the 39th International Conference on Machine Learning}, volume 162 of \emph{Proceedings of Machine Learning Research}, pp.\  3025--3039. PMLR, 17--23 Jul 2022.
\newblock URL \url{https://proceedings.mlr.press/v162/chen22a.html}.

\bibitem[Clark et~al.(2019)Clark, Lee, Chang, Kwiatkowski, Collins, and Toutanova]{clark-etal-2019-boolq}
Christopher Clark, Kenton Lee, Ming-Wei Chang, Tom Kwiatkowski, Michael Collins, and Kristina Toutanova.
\newblock {B}ool{Q}: Exploring the surprising difficulty of natural yes/no questions.
\newblock In Jill Burstein, Christy Doran, and Thamar Solorio (eds.), \emph{Proceedings of the 2019 Conference of the North {A}merican Chapter of the Association for Computational Linguistics: Human Language Technologies, Volume 1 (Long and Short Papers)}, pp.\  2924--2936, Minneapolis, Minnesota, June 2019. Association for Computational Linguistics.
\newblock \doi{10.18653/v1/N19-1300}.
\newblock URL \url{https://aclanthology.org/N19-1300}.

\bibitem[Clark et~al.(2018)Clark, Cowhey, Etzioni, Khot, Sabharwal, Schoenick, and Tafjord]{arcchallenge}
Peter Clark, Isaac Cowhey, Oren Etzioni, Tushar Khot, Ashish Sabharwal, Carissa Schoenick, and Oyvind Tafjord.
\newblock Think you have solved question answering? try arc, the {AI2} reasoning challenge.
\newblock \emph{CoRR}, abs/1803.05457, 2018.
\newblock URL \url{http://arxiv.org/abs/1803.05457}.

\bibitem[Darcet et~al.(2024)Darcet, Oquab, Mairal, and Bojanowski]{darcet2024vision}
Timoth{\'e}e Darcet, Maxime Oquab, Julien Mairal, and Piotr Bojanowski.
\newblock Vision transformers need registers.
\newblock In \emph{The Twelfth International Conference on Learning Representations}, 2024.
\newblock URL \url{https://openreview.net/forum?id=2dnO3LLiJ1}.

\bibitem[de~Jorge et~al.(2021)de~Jorge, Sanyal, Behl, Torr, Rogez, and Dokania]{jorge2021progressive}
Pau de~Jorge, Amartya Sanyal, Harkirat Behl, Philip Torr, Gr{\'e}gory Rogez, and Puneet~K. Dokania.
\newblock Progressive skeletonization: Trimming more fat from a network at initialization.
\newblock In \emph{International Conference on Learning Representations}, 2021.
\newblock URL \url{https://openreview.net/forum?id=9GsFOUyUPi}.

\bibitem[Dettmers et~al.(2022)Dettmers, Lewis, Belkada, and Zettlemoyer]{dettmers2022gptint}
Tim Dettmers, Mike Lewis, Younes Belkada, and Luke Zettlemoyer.
\newblock {GPT}3.int8(): 8-bit matrix multiplication for transformers at scale.
\newblock In Alice~H. Oh, Alekh Agarwal, Danielle Belgrave, and Kyunghyun Cho (eds.), \emph{Advances in Neural Information Processing Systems}, 2022.
\newblock URL \url{https://openreview.net/forum?id=dXiGWqBoxaD}.

\bibitem[Devlin et~al.(2019)Devlin, Chang, Lee, and Toutanova]{devlin-etal-2019-bert}
Jacob Devlin, Ming-Wei Chang, Kenton Lee, and Kristina Toutanova.
\newblock {BERT}: Pre-training of deep bidirectional transformers for language understanding.
\newblock In Jill Burstein, Christy Doran, and Thamar Solorio (eds.), \emph{Proceedings of the 2019 Conference of the North {A}merican Chapter of the Association for Computational Linguistics: Human Language Technologies, Volume 1 (Long and Short Papers)}, pp.\  4171--4186, Minneapolis, Minnesota, June 2019. Association for Computational Linguistics.
\newblock \doi{10.18653/v1/N19-1423}.
\newblock URL \url{https://aclanthology.org/N19-1423}.

\bibitem[Dubey et~al.(2024)Dubey, Jauhri, Pandey, Kadian, Al-Dahle, Letman, Mathur, Schelten, Yang, Fan, Goyal, Hartshorn, Yang, Mitra, Sravankumar, Korenev, Hinsvark, Rao, Zhang, Rodriguez, Gregerson, Spataru, Roziere, Biron, Tang, Chern, Caucheteux, Nayak, Bi, Marra, McConnell, Keller, Touret, Wu, Wong, Ferrer, Nikolaidis, Allonsius, Song, Pintz, Livshits, Esiobu, Choudhary, Mahajan, Garcia-Olano, Perino, Hupkes, Lakomkin, AlBadawy, Lobanova, Dinan, Smith, Radenovic, Zhang, Synnaeve, Lee, Anderson, Nail, Mialon, Pang, Cucurell, Nguyen, Korevaar, Xu, Touvron, Zarov, Ibarra, Kloumann, Misra, Evtimov, Copet, Lee, Geffert, Vranes, Park, Mahadeokar, Shah, van~der Linde, Billock, Hong, Lee, Fu, Chi, Huang, Liu, Wang, Yu, Bitton, Spisak, Park, Rocca, Johnstun, Saxe, Jia, Alwala, Upasani, Plawiak, Li, Heafield, Stone, El-Arini, Iyer, Malik, Chiu, Bhalla, Rantala-Yeary, van~der Maaten, Chen, Tan, Jenkins, Martin, Madaan, Malo, Blecher, Landzaat, de~Oliveira, Muzzi, Pasupuleti, Singh, Paluri, Kardas, Oldham, Rita, Pavlova, Kambadur, Lewis, Si, Singh, Hassan, Goyal, Torabi, Bashlykov, Bogoychev, Chatterji, Duchenne, Çelebi, Alrassy, Zhang, Li, Vasic, Weng, Bhargava, Dubal, Krishnan, Koura, Xu, He, Dong, Srinivasan, Ganapathy, Calderer, Cabral, Stojnic, Raileanu, Girdhar, Patel, Sauvestre, Polidoro, Sumbaly, Taylor, Silva, Hou, Wang, Hosseini, Chennabasappa, Singh, Bell, Kim, Edunov, Nie, Narang, Raparthy, Shen, Wan, Bhosale, Zhang, Vandenhende, Batra, Whitman, Sootla, Collot, Gururangan, Borodinsky, Herman, Fowler, Sheasha, Georgiou, Scialom, Speckbacher, Mihaylov, Xiao, Karn, Goswami, Gupta, Ramanathan, Kerkez, Gonguet, Do, Vogeti, Petrovic, Chu, Xiong, Fu, Meers, Martinet, Wang, Tan, Xie, Jia, Wang, Goldschlag, Gaur, Babaei, Wen, Song, Zhang, Li, Mao, Coudert, Yan, Chen, Papakipos, Singh, Grattafiori, Jain, Kelsey, Shajnfeld, Gangidi, Victoria, Goldstand, Menon, Sharma, Boesenberg, Vaughan, Baevski, Feinstein, Kallet, Sangani, Yunus, Lupu, Alvarado, Caples, Gu, Ho, Poulton, Ryan, Ramchandani, Franco, Saraf, Chowdhury, Gabriel, Bharambe, Eisenman, Yazdan, James, Maurer, Leonhardi, Huang, Loyd, Paola, Paranjape, Liu, Wu, Ni, Hancock, Wasti, Spence, Stojkovic, Gamido, Montalvo, Parker, Burton, Mejia, Wang, Kim, Zhou, Hu, Chu, Cai, Tindal, Feichtenhofer, Civin, Beaty, Kreymer, Li, Wyatt, Adkins, Xu, Testuggine, David, Parikh, Liskovich, Foss, Wang, Le, Holland, Dowling, Jamil, Montgomery, Presani, Hahn, Wood, Brinkman, Arcaute, Dunbar, Smothers, Sun, Kreuk, Tian, Ozgenel, Caggioni, Guzmán, Kanayet, Seide, Florez, Schwarz, Badeer, Swee, Halpern, Thattai, Herman, Sizov, Guangyi, Zhang, Lakshminarayanan, Shojanazeri, Zou, Wang, Zha, Habeeb, Rudolph, Suk, Aspegren, Goldman, Damlaj, Molybog, Tufanov, Veliche, Gat, Weissman, Geboski, Kohli, Asher, Gaya, Marcus, Tang, Chan, Zhen, Reizenstein, Teboul, Zhong, Jin, Yang, Cummings, Carvill, Shepard, McPhie, Torres, Ginsburg, Wang, Wu, U, Saxena, Prasad, Khandelwal, Zand, Matosich, Veeraraghavan, Michelena, Li, Huang, Chawla, Lakhotia, Huang, Chen, Garg, A, Silva, Bell, Zhang, Guo, Yu, Moshkovich, Wehrstedt, Khabsa, Avalani, Bhatt, Tsimpoukelli, Mankus, Hasson, Lennie, Reso, Groshev, Naumov, Lathi, Keneally, Seltzer, Valko, Restrepo, Patel, Vyatskov, Samvelyan, Clark, Macey, Wang, Hermoso, Metanat, Rastegari, Bansal, Santhanam, Parks, White, Bawa, Singhal, Egebo, Usunier, Laptev, Dong, Zhang, Cheng, Chernoguz, Hart, Salpekar, Kalinli, Kent, Parekh, Saab, Balaji, Rittner, Bontrager, Roux, Dollar, Zvyagina, Ratanchandani, Yuvraj, Liang, Alao, Rodriguez, Ayub, Murthy, Nayani, Mitra, Li, Hogan, Battey, Wang, Maheswari, Howes, Rinott, Bondu, Datta, Chugh, Hunt, Dhillon, Sidorov, Pan, Verma, Yamamoto, Ramaswamy, Lindsay, Lindsay, Feng, Lin, Zha, Shankar, Zhang, Zhang, Wang, Agarwal, Sajuyigbe, Chintala, Max, Chen, Kehoe, Satterfield, Govindaprasad, Gupta, Cho, Virk, Subramanian, Choudhury, Goldman, Remez, Glaser, Best, Kohler, Robinson, Li, Zhang, Matthews, Chou, Shaked, Vontimitta, Ajayi, Montanez, Mohan, Kumar, Mangla, Albiero, Ionescu, Poenaru, Mihailescu, Ivanov, Li, Wang, Jiang, Bouaziz, Constable, Tang, Wang, Wu, Wang, Xia, Wu, Gao, Chen, Hu, Jia, Qi, Li, Zhang, Zhang, Adi, Nam, Yu, Wang, Hao, Qian, He, Rait, DeVito, Rosnbrick, Wen, Yang, and Zhao]{dubey2024llama3herdmodels}
Abhimanyu Dubey, Abhinav Jauhri, Abhinav Pandey, Abhishek Kadian, Ahmad Al-Dahle, Aiesha Letman, Akhil Mathur, Alan Schelten, Amy Yang, Angela Fan, Anirudh Goyal, Anthony Hartshorn, Aobo Yang, Archi Mitra, Archie Sravankumar, Artem Korenev, Arthur Hinsvark, Arun Rao, Aston Zhang, Aurelien Rodriguez, Austen Gregerson, Ava Spataru, Baptiste Roziere, Bethany Biron, Binh Tang, Bobbie Chern, Charlotte Caucheteux, Chaya Nayak, Chloe Bi, Chris Marra, Chris McConnell, Christian Keller, Christophe Touret, Chunyang Wu, Corinne Wong, Cristian~Canton Ferrer, Cyrus Nikolaidis, Damien Allonsius, Daniel Song, Danielle Pintz, Danny Livshits, David Esiobu, Dhruv Choudhary, Dhruv Mahajan, Diego Garcia-Olano, Diego Perino, Dieuwke Hupkes, Egor Lakomkin, Ehab AlBadawy, Elina Lobanova, Emily Dinan, Eric~Michael Smith, Filip Radenovic, Frank Zhang, Gabriel Synnaeve, Gabrielle Lee, Georgia~Lewis Anderson, Graeme Nail, Gregoire Mialon, Guan Pang, Guillem Cucurell, Hailey Nguyen, Hannah Korevaar, Hu~Xu, Hugo Touvron, Iliyan Zarov, Imanol~Arrieta Ibarra, Isabel Kloumann, Ishan Misra, Ivan Evtimov, Jade Copet, Jaewon Lee, Jan Geffert, Jana Vranes, Jason Park, Jay Mahadeokar, Jeet Shah, Jelmer van~der Linde, Jennifer Billock, Jenny Hong, Jenya Lee, Jeremy Fu, Jianfeng Chi, Jianyu Huang, Jiawen Liu, Jie Wang, Jiecao Yu, Joanna Bitton, Joe Spisak, Jongsoo Park, Joseph Rocca, Joshua Johnstun, Joshua Saxe, Junteng Jia, Kalyan~Vasuden Alwala, Kartikeya Upasani, Kate Plawiak, Ke~Li, Kenneth Heafield, Kevin Stone, Khalid El-Arini, Krithika Iyer, Kshitiz Malik, Kuenley Chiu, Kunal Bhalla, Lauren Rantala-Yeary, Laurens van~der Maaten, Lawrence Chen, Liang Tan, Liz Jenkins, Louis Martin, Lovish Madaan, Lubo Malo, Lukas Blecher, Lukas Landzaat, Luke de~Oliveira, Madeline Muzzi, Mahesh Pasupuleti, Mannat Singh, Manohar Paluri, Marcin Kardas, Mathew Oldham, Mathieu Rita, Maya Pavlova, Melanie Kambadur, Mike Lewis, Min Si, Mitesh~Kumar Singh, Mona Hassan, Naman Goyal, Narjes Torabi, Nikolay Bashlykov, Nikolay Bogoychev, Niladri Chatterji, Olivier Duchenne, Onur Çelebi, Patrick Alrassy, Pengchuan Zhang, Pengwei Li, Petar Vasic, Peter Weng, Prajjwal Bhargava, Pratik Dubal, Praveen Krishnan, Punit~Singh Koura, Puxin Xu, Qing He, Qingxiao Dong, Ragavan Srinivasan, Raj Ganapathy, Ramon Calderer, Ricardo~Silveira Cabral, Robert Stojnic, Roberta Raileanu, Rohit Girdhar, Rohit Patel, Romain Sauvestre, Ronnie Polidoro, Roshan Sumbaly, Ross Taylor, Ruan Silva, Rui Hou, Rui Wang, Saghar Hosseini, Sahana Chennabasappa, Sanjay Singh, Sean Bell, Seohyun~Sonia Kim, Sergey Edunov, Shaoliang Nie, Sharan Narang, Sharath Raparthy, Sheng Shen, Shengye Wan, Shruti Bhosale, Shun Zhang, Simon Vandenhende, Soumya Batra, Spencer Whitman, Sten Sootla, Stephane Collot, Suchin Gururangan, Sydney Borodinsky, Tamar Herman, Tara Fowler, Tarek Sheasha, Thomas Georgiou, Thomas Scialom, Tobias Speckbacher, Todor Mihaylov, Tong Xiao, Ujjwal Karn, Vedanuj Goswami, Vibhor Gupta, Vignesh Ramanathan, Viktor Kerkez, Vincent Gonguet, Virginie Do, Vish Vogeti, Vladan Petrovic, Weiwei Chu, Wenhan Xiong, Wenyin Fu, Whitney Meers, Xavier Martinet, Xiaodong Wang, Xiaoqing~Ellen Tan, Xinfeng Xie, Xuchao Jia, Xuewei Wang, Yaelle Goldschlag, Yashesh Gaur, Yasmine Babaei, Yi~Wen, Yiwen Song, Yuchen Zhang, Yue Li, Yuning Mao, Zacharie~Delpierre Coudert, Zheng Yan, Zhengxing Chen, Zoe Papakipos, Aaditya Singh, Aaron Grattafiori, Abha Jain, Adam Kelsey, Adam Shajnfeld, Adithya Gangidi, Adolfo Victoria, Ahuva Goldstand, Ajay Menon, Ajay Sharma, Alex Boesenberg, Alex Vaughan, Alexei Baevski, Allie Feinstein, Amanda Kallet, Amit Sangani, Anam Yunus, Andrei Lupu, Andres Alvarado, Andrew Caples, Andrew Gu, Andrew Ho, Andrew Poulton, Andrew Ryan, Ankit Ramchandani, Annie Franco, Aparajita Saraf, Arkabandhu Chowdhury, Ashley Gabriel, Ashwin Bharambe, Assaf Eisenman, Azadeh Yazdan, Beau James, Ben Maurer, Benjamin Leonhardi, Bernie Huang, Beth Loyd, Beto~De Paola, Bhargavi Paranjape, Bing Liu, Bo~Wu, Boyu Ni, Braden Hancock, Bram Wasti, Brandon Spence, Brani Stojkovic, Brian Gamido, Britt Montalvo, Carl Parker, Carly Burton, Catalina Mejia, Changhan Wang, Changkyu Kim, Chao Zhou, Chester Hu, Ching-Hsiang Chu, Chris Cai, Chris Tindal, Christoph Feichtenhofer, Damon Civin, Dana Beaty, Daniel Kreymer, Daniel Li, Danny Wyatt, David Adkins, David Xu, Davide Testuggine, Delia David, Devi Parikh, Diana Liskovich, Didem Foss, Dingkang Wang, Duc Le, Dustin Holland, Edward Dowling, Eissa Jamil, Elaine Montgomery, Eleonora Presani, Emily Hahn, Emily Wood, Erik Brinkman, Esteban Arcaute, Evan Dunbar, Evan Smothers, Fei Sun, Felix Kreuk, Feng Tian, Firat Ozgenel, Francesco Caggioni, Francisco Guzmán, Frank Kanayet, Frank Seide, Gabriela~Medina Florez, Gabriella Schwarz, Gada Badeer, Georgia Swee, Gil Halpern, Govind Thattai, Grant Herman, Grigory Sizov, Guangyi, Zhang, Guna Lakshminarayanan, Hamid Shojanazeri, Han Zou, Hannah Wang, Hanwen Zha, Haroun Habeeb, Harrison Rudolph, Helen Suk, Henry Aspegren, Hunter Goldman, Ibrahim Damlaj, Igor Molybog, Igor Tufanov, Irina-Elena Veliche, Itai Gat, Jake Weissman, James Geboski, James Kohli, Japhet Asher, Jean-Baptiste Gaya, Jeff Marcus, Jeff Tang, Jennifer Chan, Jenny Zhen, Jeremy Reizenstein, Jeremy Teboul, Jessica Zhong, Jian Jin, Jingyi Yang, Joe Cummings, Jon Carvill, Jon Shepard, Jonathan McPhie, Jonathan Torres, Josh Ginsburg, Junjie Wang, Kai Wu, Kam~Hou U, Karan Saxena, Karthik Prasad, Kartikay Khandelwal, Katayoun Zand, Kathy Matosich, Kaushik Veeraraghavan, Kelly Michelena, Keqian Li, Kun Huang, Kunal Chawla, Kushal Lakhotia, Kyle Huang, Lailin Chen, Lakshya Garg, Lavender A, Leandro Silva, Lee Bell, Lei Zhang, Liangpeng Guo, Licheng Yu, Liron Moshkovich, Luca Wehrstedt, Madian Khabsa, Manav Avalani, Manish Bhatt, Maria Tsimpoukelli, Martynas Mankus, Matan Hasson, Matthew Lennie, Matthias Reso, Maxim Groshev, Maxim Naumov, Maya Lathi, Meghan Keneally, Michael~L. Seltzer, Michal Valko, Michelle Restrepo, Mihir Patel, Mik Vyatskov, Mikayel Samvelyan, Mike Clark, Mike Macey, Mike Wang, Miquel~Jubert Hermoso, Mo~Metanat, Mohammad Rastegari, Munish Bansal, Nandhini Santhanam, Natascha Parks, Natasha White, Navyata Bawa, Nayan Singhal, Nick Egebo, Nicolas Usunier, Nikolay~Pavlovich Laptev, Ning Dong, Ning Zhang, Norman Cheng, Oleg Chernoguz, Olivia Hart, Omkar Salpekar, Ozlem Kalinli, Parkin Kent, Parth Parekh, Paul Saab, Pavan Balaji, Pedro Rittner, Philip Bontrager, Pierre Roux, Piotr Dollar, Polina Zvyagina, Prashant Ratanchandani, Pritish Yuvraj, Qian Liang, Rachad Alao, Rachel Rodriguez, Rafi Ayub, Raghotham Murthy, Raghu Nayani, Rahul Mitra, Raymond Li, Rebekkah Hogan, Robin Battey, Rocky Wang, Rohan Maheswari, Russ Howes, Ruty Rinott, Sai~Jayesh Bondu, Samyak Datta, Sara Chugh, Sara Hunt, Sargun Dhillon, Sasha Sidorov, Satadru Pan, Saurabh Verma, Seiji Yamamoto, Sharadh Ramaswamy, Shaun Lindsay, Shaun Lindsay, Sheng Feng, Shenghao Lin, Shengxin~Cindy Zha, Shiva Shankar, Shuqiang Zhang, Shuqiang Zhang, Sinong Wang, Sneha Agarwal, Soji Sajuyigbe, Soumith Chintala, Stephanie Max, Stephen Chen, Steve Kehoe, Steve Satterfield, Sudarshan Govindaprasad, Sumit Gupta, Sungmin Cho, Sunny Virk, Suraj Subramanian, Sy~Choudhury, Sydney Goldman, Tal Remez, Tamar Glaser, Tamara Best, Thilo Kohler, Thomas Robinson, Tianhe Li, Tianjun Zhang, Tim Matthews, Timothy Chou, Tzook Shaked, Varun Vontimitta, Victoria Ajayi, Victoria Montanez, Vijai Mohan, Vinay~Satish Kumar, Vishal Mangla, Vítor Albiero, Vlad Ionescu, Vlad Poenaru, Vlad~Tiberiu Mihailescu, Vladimir Ivanov, Wei Li, Wenchen Wang, Wenwen Jiang, Wes Bouaziz, Will Constable, Xiaocheng Tang, Xiaofang Wang, Xiaojian Wu, Xiaolan Wang, Xide Xia, Xilun Wu, Xinbo Gao, Yanjun Chen, Ye~Hu, Ye~Jia, Ye~Qi, Yenda Li, Yilin Zhang, Ying Zhang, Yossi Adi, Youngjin Nam, Yu, Wang, Yuchen Hao, Yundi Qian, Yuzi He, Zach Rait, Zachary DeVito, Zef Rosnbrick, Zhaoduo Wen, Zhenyu Yang, and Zhiwei Zhao.
\newblock The llama 3 herd of models, 2024.
\newblock URL \url{https://arxiv.org/abs/2407.21783}.

\bibitem[Egiazarian et~al.(2024)Egiazarian, Panferov, Kuznedelev, Frantar, Babenko, and Alistarh]{egiazarian2024extreme}
Vage Egiazarian, Andrei Panferov, Denis Kuznedelev, Elias Frantar, Artem Babenko, and Dan Alistarh.
\newblock Extreme compression of large language models via additive quantization.
\newblock In \emph{Forty-first International Conference on Machine Learning}, 2024.
\newblock URL \url{https://openreview.net/forum?id=5mCaITRTmO}.

\bibitem[Evci et~al.(2020)Evci, Gale, Menick, Castro, and Elsen]{pmlr-v119-evci20a}
Utku Evci, Trevor Gale, Jacob Menick, Pablo~Samuel Castro, and Erich Elsen.
\newblock Rigging the lottery: Making all tickets winners.
\newblock In Hal~Daumé III and Aarti Singh (eds.), \emph{Proceedings of the 37th International Conference on Machine Learning}, volume 119 of \emph{Proceedings of Machine Learning Research}, pp.\  2943--2952. PMLR, 13--18 Jul 2020.
\newblock URL \url{https://proceedings.mlr.press/v119/evci20a.html}.

\bibitem[Frantar \& Alistarh(2023)Frantar and Alistarh]{frantar-sparsegpt}
Elias Frantar and Dan Alistarh.
\newblock {SparseGPT}: Massive language models can be accurately pruned in one-shot.
\newblock \emph{arXiv preprint arXiv:2301.00774}, 2023.

\bibitem[Gao et~al.(2024)Gao, Tow, Abbasi, Biderman, Black, DiPofi, Foster, Golding, Hsu, Le~Noac'h, Li, McDonell, Muennighoff, Ociepa, Phang, Reynolds, Schoelkopf, Skowron, Sutawika, Tang, Thite, Wang, Wang, and Zou]{eval-harness}
Leo Gao, Jonathan Tow, Baber Abbasi, Stella Biderman, Sid Black, Anthony DiPofi, Charles Foster, Laurence Golding, Jeffrey Hsu, Alain Le~Noac'h, Haonan Li, Kyle McDonell, Niklas Muennighoff, Chris Ociepa, Jason Phang, Laria Reynolds, Hailey Schoelkopf, Aviya Skowron, Lintang Sutawika, Eric Tang, Anish Thite, Ben Wang, Kevin Wang, and Andy Zou.
\newblock A framework for few-shot language model evaluation, 07 2024.
\newblock URL \url{https://zenodo.org/records/12608602}.

\bibitem[Gil(2021)]{gil2021vit}
Jacob Gil.
\newblock Vision transformer explainability.
\newblock \url{https://jacobgil.github.io/deeplearning/vision-transformer-explainability#attention-rollout}, 2021.
\newblock Accessed: 2024-09-18.

\bibitem[Hassibi \& Stork(1992)Hassibi and Stork]{NIPS1992_303ed4c6}
Babak Hassibi and David Stork.
\newblock Second order derivatives for network pruning: Optimal brain surgeon.
\newblock In S.~Hanson, J.~Cowan, and C.~Giles (eds.), \emph{Advances in Neural Information Processing Systems}, volume~5. Morgan-Kaufmann, 1992.
\newblock URL \url{https://proceedings.neurips.cc/paper_files/paper/1992/file/303ed4c69846ab36c2904d3ba8573050-Paper.pdf}.

\bibitem[Hendrycks et~al.(2021)Hendrycks, Burns, Basart, Zou, Mazeika, Song, and Steinhardt]{hendrycks2021measuring}
Dan Hendrycks, Collin Burns, Steven Basart, Andy Zou, Mantas Mazeika, Dawn Song, and Jacob Steinhardt.
\newblock Measuring massive multitask language understanding.
\newblock In \emph{International Conference on Learning Representations}, 2021.
\newblock URL \url{https://openreview.net/forum?id=d7KBjmI3GmQ}.

\bibitem[Hu et~al.(2022)Hu, yelong shen, Wallis, Allen-Zhu, Li, Wang, Wang, and Chen]{hu2022lora}
Edward~J Hu, yelong shen, Phillip Wallis, Zeyuan Allen-Zhu, Yuanzhi Li, Shean Wang, Lu~Wang, and Weizhu Chen.
\newblock Lo{RA}: Low-rank adaptation of large language models.
\newblock In \emph{International Conference on Learning Representations}, 2022.
\newblock URL \url{https://openreview.net/forum?id=nZeVKeeFYf9}.

\bibitem[Huber(1981)]{HuberRob}
P.J. Huber.
\newblock \emph{{Robust statistics}}.
\newblock Wiley New York, 1981.

\bibitem[Kovaleva et~al.(2021)Kovaleva, Kulshreshtha, Rogers, and Rumshisky]{kovaleva-etal-2021-bert}
Olga Kovaleva, Saurabh Kulshreshtha, Anna Rogers, and Anna Rumshisky.
\newblock {BERT} busters: Outlier dimensions that disrupt transformers.
\newblock In Chengqing Zong, Fei Xia, Wenjie Li, and Roberto Navigli (eds.), \emph{Findings of the Association for Computational Linguistics: ACL-IJCNLP 2021}, pp.\  3392--3405, Online, August 2021. Association for Computational Linguistics.
\newblock \doi{10.18653/v1/2021.findings-acl.300}.
\newblock URL \url{https://aclanthology.org/2021.findings-acl.300}.

\bibitem[LeCun et~al.(1989)LeCun, Denker, and Solla]{NIPS1989_6c9882bb}
Yann LeCun, John Denker, and Sara Solla.
\newblock Optimal brain damage.
\newblock In D.~Touretzky (ed.), \emph{Advances in Neural Information Processing Systems}, volume~2. Morgan-Kaufmann, 1989.
\newblock URL \url{https://proceedings.neurips.cc/paper_files/paper/1989/file/6c9882bbac1c7093bd25041881277658-Paper.pdf}.

\bibitem[Lee et~al.(2019)Lee, Ajanthan, and Torr]{lee2018snip}
Namhoon Lee, Thalaiyasingam Ajanthan, and Philip Torr.
\newblock Snip: Single-shot network pruning based on connection sensitivity.
\newblock In \emph{International Conference on Learning Representations}, 2019.
\newblock URL \url{https://openreview.net/forum?id=B1VZqjAcYX}.

\bibitem[Li et~al.(2024)Li, Tang, and Zhang]{li2024lorap}
Guangyan Li, Yongqiang Tang, and Wensheng Zhang.
\newblock Lo{RAP}: Transformer sub-layers deserve differentiated structured compression for large language models.
\newblock In \emph{Forty-first International Conference on Machine Learning}, 2024.
\newblock URL \url{https://openreview.net/forum?id=mhI5nc5QwX}.

\bibitem[Li et~al.(2023)Li, Yu, Zhang, Liang, He, Chen, and Zhao]{losparse}
Yixiao Li, Yifan Yu, Qingru Zhang, Chen Liang, Pengcheng He, Weizhu Chen, and Tuo Zhao.
\newblock {L}o{S}parse: Structured compression of large language models based on low-rank and sparse approximation.
\newblock In Andreas Krause, Emma Brunskill, Kyunghyun Cho, Barbara Engelhardt, Sivan Sabato, and Jonathan Scarlett (eds.), \emph{Proceedings of the 40th International Conference on Machine Learning}, volume 202 of \emph{Proceedings of Machine Learning Research}, pp.\  20336--20350. PMLR, 23--29 Jul 2023.
\newblock URL \url{https://proceedings.mlr.press/v202/li23ap.html}.

\bibitem[Lin et~al.(2014)Lin, Maire, Belongie, Hays, Perona, Ramanan, Doll{\'a}r, and Zitnick]{coco}
Tsung-Yi Lin, Michael Maire, Serge Belongie, James Hays, Pietro Perona, Deva Ramanan, Piotr Doll{\'a}r, and C.~Lawrence Zitnick.
\newblock Microsoft coco: Common objects in context.
\newblock In David Fleet, Tomas Pajdla, Bernt Schiele, and Tinne Tuytelaars (eds.), \emph{Computer Vision -- ECCV 2014}, pp.\  740--755, Cham, 2014. Springer International Publishing.
\newblock ISBN 978-3-319-10602-1.

\bibitem[Ma et~al.(2023)Ma, Fang, and Wang]{ma2023llmpruner}
Xinyin Ma, Gongfan Fang, and Xinchao Wang.
\newblock {LLM}-pruner: On the structural pruning of large language models.
\newblock In \emph{Thirty-seventh Conference on Neural Information Processing Systems}, 2023.
\newblock URL \url{https://openreview.net/forum?id=J8Ajf9WfXP}.

\bibitem[Mihaylov et~al.(2018)Mihaylov, Clark, Khot, and Sabharwal]{OpenBookQA2018}
Todor Mihaylov, Peter Clark, Tushar Khot, and Ashish Sabharwal.
\newblock Can a suit of armor conduct electricity? a new dataset for open book question answering.
\newblock In \emph{EMNLP}, 2018.

\bibitem[Mishra et~al.(2021)Mishra, Latorre, Pool, Stosic, Stosic, Venkatesh, Yu, and Micikevicius]{Mishra2021AcceleratingSD}
Asit~K. Mishra, Jorge~Albericio Latorre, Jeff Pool, Darko Stosic, Dusan Stosic, Ganesh Venkatesh, Chong Yu, and Paulius Micikevicius.
\newblock Accelerating sparse deep neural networks.
\newblock \emph{ArXiv}, abs/2104.08378, 2021.
\newblock URL \url{https://api.semanticscholar.org/CorpusID:233296249}.

\bibitem[Mozaffari \& Dehnavi(2024)Mozaffari and Dehnavi]{mozaffari2024slimoneshotquantizedsparse}
Mohammad Mozaffari and Maryam~Mehri Dehnavi.
\newblock Slim: One-shot quantized sparse plus low-rank approximation of llms, 2024.
\newblock URL \url{https://arxiv.org/abs/2410.09615}.

\bibitem[Mozaffari et~al.(2024)Mozaffari, Yazdanbakhsh, Zhang, and Dehnavi]{mozaffari2024slopedoubleprunedsparseplus}
Mohammad Mozaffari, Amir Yazdanbakhsh, Zhao Zhang, and Maryam~Mehri Dehnavi.
\newblock Slope: Double-pruned sparse plus lazy low-rank adapter pretraining of llms, 2024.
\newblock URL \url{https://arxiv.org/abs/2405.16325}.

\bibitem[Mozer \& Smolensky(1988)Mozer and Smolensky]{NIPS1988_07e1cd7d}
Michael~C Mozer and Paul Smolensky.
\newblock Skeletonization: A technique for trimming the fat from a network via relevance assessment.
\newblock In D.~Touretzky (ed.), \emph{Advances in Neural Information Processing Systems}, volume~1. Morgan-Kaufmann, 1988.
\newblock URL \url{https://proceedings.neurips.cc/paper_files/paper/1988/file/07e1cd7dca89a1678042477183b7ac3f-Paper.pdf}.

\bibitem[Netrapalli et~al.(2014)Netrapalli, U~N, Sanghavi, Anandkumar, and Jain]{alt_proj}
Praneeth Netrapalli, Niranjan U~N, Sujay Sanghavi, Animashree Anandkumar, and Prateek Jain.
\newblock Non-convex robust pca.
\newblock In Z.~Ghahramani, M.~Welling, C.~Cortes, N.~Lawrence, and K.Q. Weinberger (eds.), \emph{Advances in Neural Information Processing Systems}, volume~27. Curran Associates, Inc., 2014.
\newblock URL \url{https://proceedings.neurips.cc/paper_files/paper/2014/file/443cb001c138b2561a0d90720d6ce111-Paper.pdf}.

\bibitem[NeuralMagic(2021)]{deepsparse}
NeuralMagic.
\newblock Deepsparse: A cpu runtime for sparse inference of neural networks.
\newblock \url{https://github.com/neuralmagic/deepsparse}, 2021.
\newblock Accessed: 2024-09-19.

\bibitem[Oquab et~al.(2023)Oquab, Darcet, Moutakanni, Vo, Szafraniec, Khalidov, Fernandez, Haziza, Massa, El-Nouby, Howes, Huang, Xu, Sharma, Li, Galuba, Rabbat, Assran, Ballas, Synnaeve, Misra, Jegou, Mairal, Labatut, Joulin, and Bojanowski]{oquab2023dinov2}
Maxime Oquab, Timothée Darcet, Theo Moutakanni, Huy~V. Vo, Marc Szafraniec, Vasil Khalidov, Pierre Fernandez, Daniel Haziza, Francisco Massa, Alaaeldin El-Nouby, Russell Howes, Po-Yao Huang, Hu~Xu, Vasu Sharma, Shang-Wen Li, Wojciech Galuba, Mike Rabbat, Mido Assran, Nicolas Ballas, Gabriel Synnaeve, Ishan Misra, Herve Jegou, Julien Mairal, Patrick Labatut, Armand Joulin, and Piotr Bojanowski.
\newblock Dinov2: Learning robust visual features without supervision, 2023.

\bibitem[Paganini(2020)]{Paganini2020PruneR}
Michela Paganini.
\newblock Prune responsibly.
\newblock \emph{ArXiv}, abs/2009.09936, 2020.
\newblock URL \url{https://api.semanticscholar.org/CorpusID:221818901}.

\bibitem[{Qwen Team}(2024)]{qwen2.5}
{Qwen Team}.
\newblock Qwen2.5: A party of foundation models, September 2024.
\newblock URL \url{https://qwenlm.github.io/blog/qwen2.5/}.

\bibitem[Raffel et~al.(2020)Raffel, Shazeer, Roberts, Lee, Narang, Matena, Zhou, Li, and Liu]{JMLR:v21:20-074}
Colin Raffel, Noam Shazeer, Adam Roberts, Katherine Lee, Sharan Narang, Michael Matena, Yanqi Zhou, Wei Li, and Peter~J. Liu.
\newblock Exploring the limits of transfer learning with a unified text-to-text transformer.
\newblock \emph{Journal of Machine Learning Research}, 21\penalty0 (140):\penalty0 1--67, 2020.
\newblock URL \url{http://jmlr.org/papers/v21/20-074.html}.

\bibitem[Ridnik et~al.(2021)Ridnik, Ben-Baruch, Noy, and Zelnik-Manor]{ridnik2021imagenet21k}
Tal Ridnik, Emanuel Ben-Baruch, Asaf Noy, and Lihi Zelnik-Manor.
\newblock Imagenet-21k pretraining for the masses, 2021.

\bibitem[Russakovsky et~al.(2015)Russakovsky, Deng, Su, Krause, Satheesh, Ma, Huang, Karpathy, Khosla, Bernstein, Berg, and Fei-Fei]{ILSVRC15}
Olga Russakovsky, Jia Deng, Hao Su, Jonathan Krause, Sanjeev Satheesh, Sean Ma, Zhiheng Huang, Andrej Karpathy, Aditya Khosla, Michael Bernstein, Alexander~C. Berg, and Li~Fei-Fei.
\newblock {ImageNet Large Scale Visual Recognition Challenge}.
\newblock \emph{International Journal of Computer Vision (IJCV)}, 115\penalty0 (3):\penalty0 211--252, 2015.
\newblock \doi{10.1007/s11263-015-0816-y}.

\bibitem[Sakaguchi et~al.(2021)Sakaguchi, Bras, Bhagavatula, and Choi]{winogrande}
Keisuke Sakaguchi, Ronan~Le Bras, Chandra Bhagavatula, and Yejin Choi.
\newblock Winogrande: an adversarial winograd schema challenge at scale.
\newblock \emph{Commun. ACM}, 64\penalty0 (9):\penalty0 99–106, aug 2021.
\newblock ISSN 0001-0782.
\newblock \doi{10.1145/3474381}.
\newblock URL \url{https://doi.org/10.1145/3474381}.

\bibitem[Sun et~al.(2024{\natexlab{a}})Sun, Chen, Kolter, and Liu]{sun2024massive}
Mingjie Sun, Xinlei Chen, J~Zico Kolter, and Zhuang Liu.
\newblock Massive activations in large language models.
\newblock In \emph{First Conference on Language Modeling}, 2024{\natexlab{a}}.
\newblock URL \url{https://openreview.net/forum?id=F7aAhfitX6}.

\bibitem[Sun et~al.(2024{\natexlab{b}})Sun, Liu, Bair, and Kolter]{sun2024a}
Mingjie Sun, Zhuang Liu, Anna Bair, and J~Zico Kolter.
\newblock A simple and effective pruning approach for large language models.
\newblock In \emph{The Twelfth International Conference on Learning Representations}, 2024{\natexlab{b}}.
\newblock URL \url{https://openreview.net/forum?id=PxoFut3dWW}.

\bibitem[Tanaka et~al.(2020)Tanaka, Kunin, Yamins, and Ganguli]{NEURIPS2020_46a4378f}
Hidenori Tanaka, Daniel Kunin, Daniel~L Yamins, and Surya Ganguli.
\newblock Pruning neural networks without any data by iteratively conserving synaptic flow.
\newblock In H.~Larochelle, M.~Ranzato, R.~Hadsell, M.F. Balcan, and H.~Lin (eds.), \emph{Advances in Neural Information Processing Systems}, volume~33, pp.\  6377--6389. Curran Associates, Inc., 2020.
\newblock URL \url{https://proceedings.neurips.cc/paper_files/paper/2020/file/46a4378f835dc8040c8057beb6a2da52-Paper.pdf}.

\bibitem[Tong et~al.(2021)Tong, Ma, and Chi]{scale_rpca}
Tian Tong, Cong Ma, and Yuejie Chi.
\newblock Accelerating ill-conditioned low-rank matrix estimation via scaled gradient descent.
\newblock \emph{J. Mach. Learn. Res.}, 22\penalty0 (1), jan 2021.
\newblock ISSN 1532-4435.

\bibitem[Wang et~al.(2018)Wang, Singh, Michael, Hill, Levy, and Bowman]{wang-etal-2018-glue}
Alex Wang, Amanpreet Singh, Julian Michael, Felix Hill, Omer Levy, and Samuel Bowman.
\newblock {GLUE}: A multi-task benchmark and analysis platform for natural language understanding.
\newblock In Tal Linzen, Grzegorz Chrupa{\l}a, and Afra Alishahi (eds.), \emph{Proceedings of the 2018 {EMNLP} Workshop {B}lackbox{NLP}: Analyzing and Interpreting Neural Networks for {NLP}}, pp.\  353--355, Brussels, Belgium, November 2018. Association for Computational Linguistics.
\newblock \doi{10.18653/v1/W18-5446}.
\newblock URL \url{https://aclanthology.org/W18-5446}.

\bibitem[Wang et~al.(2020)Wang, Zhang, and Grosse]{Wang2020Picking}
Chaoqi Wang, Guodong Zhang, and Roger Grosse.
\newblock Picking winning tickets before training by preserving gradient flow.
\newblock In \emph{International Conference on Learning Representations}, 2020.
\newblock URL \url{https://openreview.net/forum?id=SkgsACVKPH}.

\bibitem[Wolf et~al.(2020)Wolf, Debut, Sanh, Chaumond, Delangue, Moi, Cistac, Rault, Louf, Funtowicz, Davison, Shleifer, von Platen, Ma, Jernite, Plu, Xu, Le~Scao, Gugger, Drame, Lhoest, and Rush]{wolf-etal-2020-transformers}
Thomas Wolf, Lysandre Debut, Victor Sanh, Julien Chaumond, Clement Delangue, Anthony Moi, Pierric Cistac, Tim Rault, Remi Louf, Morgan Funtowicz, Joe Davison, Sam Shleifer, Patrick von Platen, Clara Ma, Yacine Jernite, Julien Plu, Canwen Xu, Teven Le~Scao, Sylvain Gugger, Mariama Drame, Quentin Lhoest, and Alexander Rush.
\newblock Transformers: State-of-the-art natural language processing.
\newblock In Qun Liu and David Schlangen (eds.), \emph{Proceedings of the 2020 Conference on Empirical Methods in Natural Language Processing: System Demonstrations}, pp.\  38--45, Online, October 2020. Association for Computational Linguistics.
\newblock \doi{10.18653/v1/2020.emnlp-demos.6}.
\newblock URL \url{https://aclanthology.org/2020.emnlp-demos.6}.

\bibitem[Wu et~al.(2020)Wu, Xu, Dai, Wan, Zhang, Yan, Tomizuka, Gonzalez, Keutzer, and Vajda]{wu2020visual}
Bichen Wu, Chenfeng Xu, Xiaoliang Dai, Alvin Wan, Peizhao Zhang, Zhicheng Yan, Masayoshi Tomizuka, Joseph Gonzalez, Kurt Keutzer, and Peter Vajda.
\newblock Visual transformers: Token-based image representation and processing for computer vision, 2020.

\bibitem[Xia et~al.(2022)Xia, Zhong, and Chen]{xia2022structured}
Mengzhou Xia, Zexuan Zhong, and Danqi Chen.
\newblock Structured pruning learns compact and accurate models.
\newblock In \emph{Association for Computational Linguistics (ACL)}, 2022.

\bibitem[Yi et~al.(2016)Yi, Park, Chen, and Caramanis]{fast_rpca}
Xinyang Yi, Dohyung Park, Yudong Chen, and Constantine Caramanis.
\newblock Fast algorithms for robust pca via gradient descent.
\newblock In \emph{Proceedings of the 30th International Conference on Neural Information Processing Systems}, NIPS'16, pp.\  4159–4167, Red Hook, NY, USA, 2016. Curran Associates Inc.
\newblock ISBN 9781510838819.

\bibitem[Yin et~al.(2024{\natexlab{a}})Yin, Liu, Jaiswal, Kundu, and Wang]{yin2024junk}
Lu~Yin, Shiwei Liu, Ajay~Kumar Jaiswal, Souvik Kundu, and Zhangyang Wang.
\newblock Junk {DNA} hypothesis: A task-centric angle of {LLM} pre-trained weights through sparsity, 2024{\natexlab{a}}.
\newblock URL \url{https://openreview.net/forum?id=EmUVpfrXWN}.

\bibitem[Yin et~al.(2024{\natexlab{b}})Yin, Wu, Zhang, Hsieh, Wang, Jia, Li, JAISWAL, Pechenizkiy, Liang, Bendersky, Wang, and Liu]{yin2024outlier}
Lu~Yin, You Wu, Zhenyu Zhang, Cheng-Yu Hsieh, Yaqing Wang, Yiling Jia, Gen Li, AJAY~KUMAR JAISWAL, Mykola Pechenizkiy, Yi~Liang, Michael Bendersky, Zhangyang Wang, and Shiwei Liu.
\newblock Outlier weighed layerwise sparsity ({OWL}): A missing secret sauce for pruning {LLM}s to high sparsity.
\newblock In \emph{Forty-first International Conference on Machine Learning}, 2024{\natexlab{b}}.
\newblock URL \url{https://openreview.net/forum?id=ahEm3l2P6w}.

\bibitem[Yu et~al.(2022{\natexlab{a}})Yu, Huang, Wang, Cheng, Chu, and Cui]{widthdepth}
Fang Yu, Kun Huang, Meng Wang, Yuan Cheng, Wei Chu, and Li~Cui.
\newblock Width and depth pruning for vision transformers.
\newblock \emph{Proceedings of the AAAI Conference on Artificial Intelligence}, 36\penalty0 (3):\penalty0 3143--3151, Jun. 2022{\natexlab{a}}.
\newblock \doi{10.1609/aaai.v36i3.20222}.
\newblock URL \url{https://ojs.aaai.org/index.php/AAAI/article/view/20222}.

\bibitem[Yu \& Xiang(2023)Yu and Xiang]{xpruner}
Lu~Yu and Wei Xiang.
\newblock X-pruner: explainable pruning for vision transformers.
\newblock In \emph{2023 IEEE/CVF Conference on Computer Vision and Pattern Recognition (CVPR)}, pp.\  24355--24363, 2023.
\newblock \doi{10.1109/CVPR52729.2023.02333}.

\bibitem[Yu et~al.(2022{\natexlab{b}})Yu, Chen, Shen, Yuan, Tan, Yang, Liu, and Wang]{yu2022unified}
Shixing Yu, Tianlong Chen, Jiayi Shen, Huan Yuan, Jianchao Tan, Sen Yang, Ji~Liu, and Zhangyang Wang.
\newblock Unified visual transformer compression.
\newblock In \emph{International Conference on Learning Representations}, 2022{\natexlab{b}}.
\newblock URL \url{https://openreview.net/forum?id=9jsZiUgkCZP}.

\bibitem[Yu et~al.(2017)Yu, Liu, Wang, and Tao]{Yu_2017_CVPR}
Xiyu Yu, Tongliang Liu, Xinchao Wang, and Dacheng Tao.
\newblock On compressing deep models by low rank and sparse decomposition.
\newblock In \emph{Proceedings of the IEEE Conference on Computer Vision and Pattern Recognition (CVPR)}, July 2017.

\bibitem[Zellers et~al.(2019)Zellers, Holtzman, Bisk, Farhadi, and Choi]{zellers2019hellaswag}
Rowan Zellers, Ari Holtzman, Yonatan Bisk, Ali Farhadi, and Yejin Choi.
\newblock Hellaswag: Can a machine really finish your sentence?
\newblock In \emph{Proceedings of the 57th Annual Meeting of the Association for Computational Linguistics}, 2019.

\bibitem[Zhang et~al.(2024{\natexlab{a}})Zhang, Chen, Shen, Yang, Ou, Yu, and Zhuang]{zhang2024loraprune}
Mingyang Zhang, Hao Chen, Chunhua Shen, Zhen Yang, Linlin Ou, Xinyi Yu, and Bohan Zhuang.
\newblock Lo{RAP}rune: Pruning meets low-rank parameter-efficient fine-tuning, 2024{\natexlab{a}}.
\newblock URL \url{https://openreview.net/forum?id=9KVT1e1qf7}.

\bibitem[Zhang et~al.(2024{\natexlab{b}})Zhang, Zhao, Lin, Yunyun, Yao, Han, Tanner, Liu, and Ji]{zhang2024dynamic}
Yuxin Zhang, Lirui Zhao, Mingbao Lin, Sun Yunyun, Yiwu Yao, Xingjia Han, Jared Tanner, Shiwei Liu, and Rongrong Ji.
\newblock Dynamic sparse no training: Training-free fine-tuning for sparse {LLM}s.
\newblock In \emph{The Twelfth International Conference on Learning Representations}, 2024{\natexlab{b}}.
\newblock URL \url{https://openreview.net/forum?id=1ndDmZdT4g}.

\bibitem[Zhao et~al.(2024{\natexlab{a}})Zhao, Hajishirzi, and Cao]{zhao2024apt}
Bowen Zhao, Hannaneh Hajishirzi, and Qingqing Cao.
\newblock {APT}: Adaptive pruning and tuning pretrained language models for efficient training and inference.
\newblock In \emph{Forty-first International Conference on Machine Learning}, 2024{\natexlab{a}}.
\newblock URL \url{https://openreview.net/forum?id=sb81Xl50JG}.

\bibitem[Zhao et~al.(2024{\natexlab{b}})Zhao, Zhang, Chen, Wang, Anandkumar, and Tian]{zhao2024galore}
Jiawei Zhao, Zhenyu Zhang, Beidi Chen, Zhangyang Wang, Anima Anandkumar, and Yuandong Tian.
\newblock Galore: Memory-efficient {LLM} training by gradient low-rank projection.
\newblock In \emph{5th Workshop on practical ML for limited/low resource settings}, 2024{\natexlab{b}}.
\newblock URL \url{https://openreview.net/forum?id=AzqPyO22zt}.

\bibitem[Zhao et~al.(2024{\natexlab{c}})Zhao, Behnia, Vakilian, and Thrampoulidis]{zhao2024implicit}
Yize Zhao, Tina Behnia, Vala Vakilian, and Christos Thrampoulidis.
\newblock Implicit geometry of next-token prediction: From language sparsity patterns to model representations.
\newblock In \emph{First Conference on Language Modeling}, 2024{\natexlab{c}}.
\newblock URL \url{https://openreview.net/forum?id=qyilOnIRHI}.

\bibitem[Zhou \& Tao(2011)Zhou and Tao]{godec}
Tianyi Zhou and Dacheng Tao.
\newblock Godec: randomized low-rank \& sparse matrix decomposition in noisy case.
\newblock In \emph{Proceedings of the 28th International Conference on International Conference on Machine Learning}, ICML'11, pp.\  33–40, Madison, WI, USA, 2011. Omnipress.
\newblock ISBN 9781450306195.

\bibitem[Zhou et~al.(2010)Zhou, Li, Wright, Cand{\`e}s, and Ma]{spcp}
Zihan Zhou, Xiaodong Li, John Wright, Emmanuel Cand{\`e}s, and Yi~Ma.
\newblock Stable principal component pursuit.
\newblock In \emph{2010 IEEE International Symposium on Information Theory, ISIT 2010 - Proceedings}, IEEE International Symposium on Information Theory - Proceedings, pp.\  1518--1522, 2010.
\newblock ISBN 9781424469604.
\newblock \doi{10.1109/ISIT.2010.5513535}.
\newblock 2010 IEEE International Symposium on Information Theory, ISIT 2010 ; Conference date: 13-06-2010 Through 18-06-2010.

\bibitem[Zhu \& Gupta(2018)Zhu and Gupta]{zhugupta}
Michael~H. Zhu and Suyog Gupta.
\newblock To prune, or not to prune: Exploring the efficacy of pruning for model compression, 2018.
\newblock URL \url{https://openreview.net/forum?id=S1lN69AT-}.

\bibitem[Zhu et~al.(2021)Zhu, Han, Tang, and Wang]{vtp}
Mingjian Zhu, Kai Han, Yehui Tang, and Yunhe Wang.
\newblock Visual transformer pruning.
\newblock \emph{CoRR}, abs/2104.08500, 2021.
\newblock URL \url{https://arxiv.org/abs/2104.08500}.

\end{thebibliography}
\bibliographystyle{iclr2025_conference}

\appendix
\section{Appendix}
\subsection{Additional Related Works}
\paragraph{Sparse and Low-Rank Decomposition for Pruning} \citet{Yu_2017_CVPR} introduced a method for sparse and low-rank decomposition of CNNs, including AlexNet and GoogLeNet, by solving the following optimization problem:
$$ \min_{\mS, \mL \in \mathbb{R}^{d_{out}\times d_{in}}} \|\mY - (\mS+\mL)\mX\|_2^{2} \; \text{s.t.} \; \|\mW - (\mS + \mL) \|_F^2\leq \gamma, \text{Rank}(\mL)\leq r, \|\mS\|_0 \leq k $$
where $\mY = \mW \mX$. In contrast, OATS employs a different approach, solving:
$$\min_{\mS, \mL \in \mathbb{R}^{d_{out} \times d_{in}}} \| \mW - \mS - \mL \|_F^2 \; \text{ s.t. } \; \text{Rank}(\mL) \leq r, \; \|\mS\|_0 \leq k. $$
A key distinction between these methods lies in their objectives: the former directly minimizes reconstruction error, while OATS adopts a simpler formulation. One might question why not follow the approach of minimizing reconstruction error. As noted in DSNoT \citep{zhang2024dynamic}, pruning methods that prioritize minimizing reconstruction error can degrade model performance in large transformers, particularly in the presence of outlier features. Their findings highlight the importance of avoiding pruning weights within outlier channels. Since feature outliers are a phenomenon unique to large transformer models \citep{dettmers2022gptint}, this issue would not have been relevant to the work of \citet{Yu_2017_CVPR}, which predates the transformer era.
\paragraph{Pruning Algorithms for Vision Transformers} There are a number of pruning approaches that have been specifically catered towards pruning vision transformers \citep{vtp, chen2021chasing, chavan2022vision, widthdepth, yu2022unified, xpruner}. However, as much of the pruning literature developed on vision transformers involved models of much smaller scale than the large language models employed in this study, almost all of the prominent pruning algorithms require some form of training on the model parameters. As OATS was designed to require no training, OATS and the aforementioned pruning algorithms would not be comparable.
\paragraph{Low-Rank Adapters during Pre-Training} In \citet{mozaffari2024slopedoubleprunedsparseplus}, the authors propose SLOPE, a novel method for accelerating the pre-training phase of LLMs by incorporating N:M sparsity and adding low-rank components to the model weights to enhance model capacity. Similar to OATS, SLOPE leads to a sparse plus low-rank structure in the model’s weight matrices, however, the low-rank terms are introduced during the final phase of pre-training and are actively trained on the model loss function. In contrast, OATS is designed as a lightweight method to accelerate inference. OATS does not require any training or fine-tuning, but instead approximates pre-trained weight matrices by solving the Robust PCA problem.
\paragraph{Quantized Sparse Low-Rank Approximation} An independent and concurrent work with OATS proposes SLIM \citep{mozaffari2024slimoneshotquantizedsparse}, a novel pipeline that combines pruning and quantization. To restore lost performance from compression, SLIM derives a low-rank term using singular-value thresholding and adopts a scaling technique akin to OATS. However, instead of the $L^2$ norm, SLIM utilizes the average absolute value across the batch and sequence dimensions. As a further deviation from OATS, SLIM is also not performing an alternating thresholding algorithm. Instead, they perform a single quantization and pruning step to initialize the quantized and sparse terms, followed by a single singular value thresholding step to establish the low-rank term.

\subsection{Time Complexity and Wall-Clock Time for OATS}
The time complexity for OATS is $\mathcal{O}(LN\alpha)$ where $L$ is the number of transformer blocks, $N$ is number of iterations, and 
$$\alpha = \max_{\mW} d^{\mW}_{out} \cdot d^{\mW}_{in} \cdot r^{\mW}$$
where the max is taken over the weight matrices, $\mW \in \mathbb{R}^{d^{\mW}_{out} \times d^{\mW}_{in}}$, in a transformer block and $r^{\mW}$ is the rank of the low-rank term for that weight matrix. The value $\alpha$ represents the time complexity needed to perform the singular value thresholding in OATS.  

Table \ref{table:app_wall} below reports the wall-clock time needed to perform a single iteration of the alternating threshold algorithm for a single transformer block for the different models that were compressed. All experiments utilized a single NVIDIA A40 with 48GB of GPU memory.

\begin{table}[H]
\centering
\begin{tabular}{cccccc}
\toprule
\multicolumn{2}{c}{\textbf{Phi-3}} & \multicolumn{2}{c}{\textbf{Llama-3}} \\ 
\textbf{Mini ($\mathbf{3.8}$B)} & \textbf{Medium ($\mathbf{14}$B)} & \textbf{8B} & \textbf{70B}\\ 
\cmidrule(lr){1-4}
8.85 & 26.02 & 17.10 & 152.80 \\
\bottomrule
\end{tabular}
\caption{Wall-clock time (in seconds) needed to perform a single iteration of the alternating projection algorithm in OATS.} \label{table:app_wall}
\end{table}
While OATS does require more wall-clock time than prior pruning algorithms, in practice, model compression would only need to be performed once before deployment. This trade-off is therefore worthwhile given the substantial performance improvements, particularly on more challenging tasks like MMLU (see Table \ref{table:mmlu}). Furthermore, like prior pruning algorithms, compressing the layers within a single transformer block can be done in parallel. For example, the time needed per transformer block of Llama-3 70B can be reduced to 71.10 seconds by compressing in parallel across four NVIDIA A40 GPUs. 

The total wall-clock time can also be reduced by lowering the number of OATS iterations. Presented in Table \ref{table:app_20} is an exploratory experiment compressing Llama-3 70B by 50\% with a rank ratio of 0.3 with only 20 iterations. Even with only a quarter of the iterations, OATS is still able to outperform all prior pruning algorithms across all performance metrics.
\begin{table}[H]
\centering
\begin{tabular}{ccc}
\toprule
 \textbf{MMLU $(\uparrow)$} & \textbf{Zero-shot $(\uparrow)$} & \textbf{Perplexity $(\downarrow)$} \\ 
\midrule 
74.02 & 73.41 & 4.95 \\ 
\bottomrule
\end{tabular}
\caption{Exploratory experiment measuring the performance of OATS on Llama-3 70B with only 20 iterations.} \label{table:app_20}
\end{table}
\subsection{Using a Robust Scaling Matrix}
To explore whether the scaling matrix $\mD$ is truly related to the outlier information, we run the following two experiments:
\begin{itemize}
\item Scaling by the square root of the features' second moments, as is currently done in OATS.
\item Scaling by the median of the features' absolute values (computed along batch and sequence dimensions):
$$\mD_{robust} = \text{median}(|X|) $$
\end{itemize}
The second experiment estimates the square root of the second moment of features in a manner that is robust (insensitive) to outliers akin to the Median Absolute Deviation estimator from the robust statistics literature \citep{HuberRob}. The results of the two experiments are presented in Table \ref{app:rob} below:
\begin{table}[H]
\centering
\begin{tabular}{cccc}
\toprule
 \textbf{Scaling Matrix} & \textbf{MMLU $(\uparrow)$} & \textbf{Zero-shot $(\uparrow)$} & \textbf{Perplexity $(\downarrow)$} \\ 
\midrule 
$\mD_{robust}$ & 55.54 & 65.77 & 18.59  \\
\midrule
 $\mD$ & \textbf{59.99} & \textbf{68.41} & \textbf{15.18}  \\ 
\bottomrule
\end{tabular}
\caption{Results of OATS on Phi-3-Mini, at $50\%$ compression rate, with a rank ratio of $25\%$ using different scaling matrices.} 
\label{app:rob}
\end{table}
The findings show that using the robust scaling method results in significantly worse performance. Hence, the scaling matrix $\mD$ that is sensitive to the outlier features and captures their scale leads to better compression.

\subsection{Switching the Order of Thresholding}
OATS opts to perform the singular-value thresholding first followed by the hard thresholding similar to \citet{godec}. However, one might consider whether the alternative order could lead to faster convergence or a better approximation. Presented in Table \ref{app:switch} below is an extension of the ablation studies presented in Section \ref{sec:hyper}, reporting the performance of OATS where the hard-thresholding is performed first:
\begin{table}[H]
\centering
\begin{tabular}{lccc}
\toprule
 \textbf{First Thresholding Operation} & \textbf{MMLU $(\uparrow)$} & \textbf{Zero-shot $(\uparrow)$} & \textbf{Perplexity $(\downarrow)$} \\ 
\midrule
Hard-Thresholding & 65.51 & 70.54 & 11.72  \\ 
\midrule
Singular Value Thresholding (OATS)  & \textbf{65.84} & \textbf{70.71} & \textbf{11.50}  \\ 
\bottomrule
\end{tabular}
\caption{Ablation results of switching of the order between the two thresholding operations. Experiments were run on Phi-3-Mini, at $40\%$ compression rate, with a rank ratio of $20\%$.} \label{app:switch}
\end{table}
While the performance still remains competitive, across all performance metrics, the switched order falls short of matching the original order presented in Algorithm \ref{alg:altmin}. 
\subsection{Magnitude-Based Pruning for the Sparse Component}
Another question that we explored is whether it is sufficient to capture the outlier information entirely in the low-rank term and determine the sparse term through a hard-thresholding that does not depend on the scaling:
$$\mS = \textsc{HardThreshold}((\mW \mD - \mL) \mD^{-1}, k).$$
Presented in Table \ref{app:mag_prune} below are the results:
\begin{table}[H]
\centering
\begin{tabular}{lccc}
\toprule
 \textbf{Outlier Scaling} & \textbf{MMLU $(\uparrow)$} & \textbf{Zero-shot $(\uparrow)$} & \textbf{Perplexity $(\downarrow)$} \\ 
\midrule 
Low-Rank Term Only  & 65.22 & \textbf{71.01} & 12.49  \\
Both Terms (OATS)  & \textbf{65.84} & 70.71 & \textbf{11.50}  \\ 
\bottomrule
\end{tabular}
\caption{Ablation results of OATS on Phi-3-Mini, at $40\%$ compression rate, with a rank ratio of $20\%$ testing whether the outlier information can be entirely captured by the low-rank term.} \label{app:mag_prune}
\end{table}
\subsection{Additional CPU Speedup Experiments}\label{app:cpu_speedup_2}
We provide additional experiments measuring the CPU speedup of OATS compared to dense models. Following the same setup as Section \ref{sec:cpu_speedup}, we run end-to-end inference on a compressed Phi-3 Medium 15B model, \textit{but instead with a sequence consisting of 256 tokens}, on an Intel Xeon Gold 6148 CPU @ 2.40GHz with 32 cores. The results are presented in Table \ref{app:cpu_table} below:

\begin{table}[H]
\centering
\begin{tabular}{lccc}
\toprule
 \multirow{2}{*}{\textbf{Compression}} & \multirow{2}{*}{\textbf{Method}} & \multirow{2}{*}{\textbf{Throughput}} & \multirow{2}{*}{\textbf{Speedup}} \\ 
 & & &  \\
\midrule 
0\% & Dense &  0.37 & $1.00\times$  \\ 
\midrule
\multirow{2}{*}{30\%} & Unstructured Pruning  & 0.40 & $1.08\times$   \\
& OATS &  0.40 &   $1.08\times$ \\ 
\midrule
\multirow{2}{*}{40\%} & Unstructured Pruning &  0.43 & $1.16\times$   \\
& OATS &  0.43 &   $1.16\times$ \\ 
\midrule
\multirow{2}{*}{50\%} & Unstructured Pruning  &  0.49 & $1.32\times$   \\
& OATS &  0.49 &   $1.32\times$ \\%\\ \midrule
\bottomrule
\end{tabular}
\caption{Comparison of throughput (tokens/second) and speedup achieved through OATS and unstructured pruning methods relative to their dense counterparts on a sequence consisting of 256 tokens.} \label{app:cpu_table}
\end{table}
Although OATS and unstructured pruning now deliver comparable CPU speed‑ups, we expect further gains are achievable by designing optimizations that explicitly exploit OATS' sparse plus low‑rank structure.
\subsection{Additional Hyperparameter Tests for OATS}
Presented in Table \ref{app:more_hyper} below includes more hyperparameters that we experimented with for the Phi-3 Mini and Llama-3 8B models. 
\begin{table}[H]
\centering
\begin{tabular}{l|cc|ccc}
\toprule
\textbf{Model} & \textbf{Compression} & \textbf{Rank Ratio}  & \textbf{MMLU ($\uparrow$)} & \textbf{Zero-Shot ($\uparrow$)}& \textbf{Perplexity ($\downarrow$)}  \\ \hline
\multirow{9}{*}{Phi-3 Mini} & \multirow{3}{*}{30\%}  & 0.1 & 68.70 & 71.65 & 10.24 \\ 
                          & & 0.2& 68.02  & 71.81 & 10.21\\
                          & & 0.3& 69.28  & 72.07 & 10.28 \\\cline{2-6}
& \multirow{3}{*}{40\%}  & 0.1 & 65.75 & 69.94 & 11.57\\
                        & & 0.2  & 65.84 & 70.71 & 11.50 \\ 
                        & & 0.3& 66.81 & 70.54 & 11.60 \\\cline{2-6}
& \multirow{3}{*}{50\%}    & 0.1  & 57.96 & 67.37 & 15.48\\ 
                          & & 0.2  & 59.12 & 68.02 & 15.13 \\ 
                          & & 0.3& 58.68 & 68.63 & 15.47 \\\hline
\multirow{6}{*}{Llama-3 8B} & \multirow{2}{*}{30\%}  & 0.1  &63.62 &68.99 & 9.35\\ 
                          & & 0.2  & 63.09 & 69.54&9.09  \\ \cline{2-6}
& \multirow{2}{*}{40\%}  &  0.1& 61.44 & 68.23 &9.23 \\
                        & & 0.2  & 61.97  &68.43 &9.09 \\ \cline{2-6}
& \multirow{2}{*}{50\%}    & 0.1  & 56.46 & 65.33 &10.85 \\ 
                          & & 0.2  & 56.07 &65.51 &10.70 \\
\bottomrule
\end{tabular}
\caption{Further experiments testing different hyperparameter configurations for OATS on the Phi-3 Mini and Llama-3 8B models.} \label{app:more_hyper}
\end{table}
\subsection{Performance Gap Between OATS and Wanda}
To better understand the increase in performance induced by the addition of the low-rank term in OATS, we have compiled in Table \ref{app:wanda_delta} below the performance gaps between OATS and Wanda.
\begin{table}[H]
\centering
\begin{tabular}{l|cccc}
\toprule
\textbf{Model} & \textbf{Compression}  & \textbf{MMLU ($\uparrow$)} & \textbf{Zero-Shot ($\uparrow$)}& \textbf{Perplexity ($\downarrow$)}  \\ \hline
\multirow{3}{*}{Phi-3 Mini} & 30\% & +1.21\% & +0.82\% & -0.44 \\ 
& 40\%   & +1.60\% & +1.24\% & -1.06\\ 
& 50\%     & +5.42\% & +3.38\% & -2.05\\ \hline
\multirow{3}{*}{Phi-3 Medium} & 30\% & +0.97\% & -0.01\% & -0.43 \\ 
& 40\%   & +1.65\% & +1.45\% & -0.79\\ 
& 50\%     & +2.52\% & +2.43\% & -1.07\\ \hline
\multirow{3}{*}{Llama-3 8B}& 30\% & +1.55\% & +0.71\% & +0.20 \\ 
& 40\%   & +2.13\% & +1.64\% & -0.50\\ 
& 50\%     & +6.63\% & +2.44\% & -1.49\\ \hline
\multirow{3}{*}{Llama-3 70B}& 30\% & -0.68\% & +0.05\% & -0.17 \\ 
& 40\%   & +0.73\% & +0.78\% & -0.40\\ 
& 50\%     & +2.74\% & +0.45\% & -0.60\\ 
\bottomrule
\end{tabular}
\caption{The impact of including a low-rank term in OATS compared to Wanda.} \label{app:wanda_delta}
\end{table}
\subsection{Qwen 2.5 Experiments}\label{app:qwen25}
Presented in Table \ref{table:qwen25} below are additional experiments benchmarking OATS against prior pruning algorithms on the Qwen 2.5 3B Instruct model \citep{qwen2.5}. All OATS experiments utilize a rank ratio of $0.2$ and $80$ iterations. 
\begin{table}[H]
\centering
\begin{tabular}{lcccc}
\toprule
  \textbf{Compression} & \textbf{Method}  & \textbf{MMLU ($\uparrow$)} & \textbf{Zero-Shot ($\uparrow$)}& \textbf{Perplexity ($\downarrow$)} \\
\midrule 
$0\%$ & Dense & 65.99 & 68.49 & 11.02 \\
\midrule
 \multirow{4}{*}{30\%} & SparseGPT  & \textbf{65.65} & 67.91 & 11.55 \\
 &Wanda     &  65.46 & 68.08 & 11.66 \\ 
 &DSNoT   & 65.65 & 68.21 & 11.67 \\
 &\algname & 65.36 & \textbf{68.74} & \textbf{11.45} \\
 \midrule
 \multirow{4}{*}{40\%} & SparseGPT  & 63.04 & 67.64 & 12.56  \\
 &Wanda     &  61.88 & 67.14 & 12.89  \\ 
 &DSNoT   & 62.26 & 67.42 &  12.91 \\
 &\algname & \textbf{64.30} & \textbf{68.76} & \textbf{12.31} \\
 \midrule
 \multirow{4}{*}{50\%} & SparseGPT  & 57.43 & 64.36 & 14.92 \\
 &Wanda     &  55.39 & 64.10 & 16.27  \\ 
 &DSNoT   & 55.78 & 64.77 &  16.43 \\
 &\algname & \textbf{58.78} & \textbf{65.74} & \textbf{14.91}  \\
\bottomrule
\end{tabular}
\caption{Benchmarks for OATS on the Qwen 2.5 3B Instruct model.} \label{table:qwen25}
\end{table}
\subsection{MMLU Subjects}\label{app:mmlu}
We evaluate on the following MMLU subjects:
\begin{itemize}
    \item Abstract Algebra
    \item Business Ethics
    \item College Computer Science
    \item College Mathematics
    \item Conceptual Physics
    \item Formal Logic
    \item Machine Learning 
    \item Miscellaneous
    \item Philosophy
    \item Global Facts
\end{itemize}
which aligns with the subset utilized in the codebase of \citet{ashkboos2024slicegpt} that can be found here: \href{https://github.com/microsoft/TransformerCompression}{https://github.com/microsoft/TransformerCompression}.
\subsection{Attention Rollout: Details}\label{app:rollout_viz}
To generate the attention rollout visualizations depicted in Section \ref{sec:rollout}, we average the attention matrices across the attention heads and discard the bottom $40\%$ attention pixels. The act of discarding the lowest value attention pixels was inspired by the following blog post by \citet{gil2021vit}.
\subsection{Zero-Shot Task-Specific Performance}\label{app:zero_shot} 
Table \ref{table:zero_shot_specific}, below, shows the task-specific performance for the zero-shot evaluation results presented in Section \ref{sec:results_llm} and Appendix \ref{app:pruning_hyper}. 
\begin{table}[H]
\centering
\begin{adjustbox}{width=\textwidth}
\begin{tabular}{l|cccccccccc}
\toprule
\textbf{Model} & \textbf{Compression} & \textbf{Method}  & \textbf{PIQA} & \textbf{HellaSwag} & \textbf{WinoGrande} & \textbf{OpenBookQA} & \textbf{RTE} & \textbf{BoolQ} & \textbf{ARC-e} & \textbf{ARC-c} \\ \hline
\multirow{16}{*}{Phi-3 Mini} & 0\% & Dense & 81.23 & 77.50 &73.56 & 46.80 & 75.81 &85.32 &78.45 & 57.25 \\  \cline{2-11}
 & \multirow{5}{*}{30\%} & SparseGPT  & 78.94 & 76.94 & 69.85 & 49.60 & 73.29 & 84.13 & 76.39 & 55.89 \\ 
                          & & Wanda  & 79.65 & 76.27 & 71.59 & 48.00 & 73.65 & 83.70 & 77.23 & 55.20 \\ %
                          & & DSNoT w/ SparseGPT  &80.09 & 75.61 & 72.22 & 47.40 & 74.37 & 84.22 & 77.86 & 54.69 \\ %
                          & & DSNoT w/ Wanda  & 80.41 & 75.52 & 72.06 & 47.60 & 74.37 & 84.53 & 79.55 & 55.55 \\  %
                          & & OATS & 80.03 & 77.07 & 72.61 & 47.60 & 74.37 & 84.92 & 77.44 & 57.76 \\ \cline{2-11} %
& \multirow{5}{*}{40\%}  & SparseGPT  & 78.35 & 75.07 & 68.59 & 47.00 & 72.20 & 83.67 & 75.29 & 53.24 \\ 
                          & & Wanda & 78.35 & 73.87 & 69.30 & 45.40 & 71.84 & 83.18 & 76.52 & 51.96 \\ 
                          & & DSNoT w/ SparseGPT\ & 78.56 & 72.99 & 70.64 & 46.20 & 70.40 & 82.72 & 76.52 & 52.82 \\ %
                          & & DSNoT w/ Wanda  & 79.33 & 73.06 & 70.88 & 44.00 & 70.76 & 83.79 & 77.40 & 53.41 \\ %
                          & & OATS & 79.38 & 75.86 & 70.01 & 46.60 & 72.56 & 83.98 & 76.85 & 55.12 \\  \cline{2-11}
& \multirow{5}{*}{50\%}    & SparseGPT  & 77.20 & 70.63 & 66.46 & 45.20 & 70.76 & 83.06 & 70.58 & 47.01  \\ 
                          & & Wanda  & 76.33 & 67.70 & 66.38 & 41.80 & 66.43 & 81.83 & 72.43 & 47.35  \\ 
                          & & DSNoT w/ SparseGPT  & 76.28 & 67.16 & 65.90 & 42.20 & 63.90 & 81.56 & 72.90 & 48.04 \\ 
                          & & DSNoT w/ Wanda  & 75.52 & 66.54 & 67.64 & 43.00 & 65.34 & 82.54 & 73.48 & 48.55 \\ %
                          & & OATS & 77.26 & 71.64 & 69.53 & 44.80 & 73.65 & 81.28 & 77.10 & 52.05 \\ \hline
\multirow{16}{*}{Phi-3 Medium} & 0\% & Dense & 81.66 & 82.83 & 75.85 & 50.00 & 77.62 & 88.17 & 78.41 & 59.64 \\  \cline{2-11}
 & \multirow{5}{*}{30\%} & SparseGPT  & 81.39 & 82.02 & 75.77 & 50.80 & 77.26 & 87.80 & 80.05 & 61.18 \\ 
                          & & Wanda  & 81.39 & 80.88 & 76.01 & 49.40 & 76.90 & 87.74 & 79.59 & 60.49 \\ 
                          & & DSNoT w/ SparseGPT  & 81.94 & 80.76 & 76.95 & 48.40 & 75.81 & 87.65 & 79.25 & 59.81 \\
                          & & DSNoT w/ Wanda  & 81.66 & 81.03 & 77.27 & 49.20 & 76.53 & 87.80 & 78.96 & 59.81 \\
                          & & OATS  & 81.07 & 82.09 & 74.43 & 51.20 & 78.34 & 88.38 & 78.16 & 58.70 \\ \cline{2-11} %
& \multirow{5}{*}{40\%}  & SparseGPT  & 80.41 & 80.70 & 75.53 & 51.20 & 77.26 & 88.32 & 81.23 & 60.58 \\ 
                          & & Wanda  & 79.87 & 78.15 & 75.45 & 48.60 & 77.26 & 87.71 & 78.11 & 58.96 \\ 
                          & & DSNoT w/ SparseGPT  & 79.82 & 78.07 & 75.37 & 47.00 & 76.53 & 87.98 & 77.31 & 58.19 \\
                          & & DSNoT w/ Wanda  & 80.30 & 78.11 & 74.66 & 47.80 & 77.26 & 88.04 & 78.11 & 58.87 \\
                          & & OATS  & 81.39 & 81.72 & 75.06 & 51.00 & 77.62 & 87.65 & 80.39 & 60.84 \\ \cline{2-11} %
& \multirow{5}{*}{50\%}    & SparseGPT  & 79.71 & 78.27 & 73.64 & 50.40 & 75.45 & 87.09 & 82.03 & 59.39 \\ 
                          & & Wanda  & 78.29 & 74.07 & 74.03 & 45.00 & 75.81 & 85.72 & 77.44 & 57.34 \\ 
                          & & DSNoT w/ SparseGPT  & 79.27 & 74.30 & 74.59 & 44.40 & 76.90 & 85.26 & 77.69 & 56.57 \\
                          & & DSNoT w/ Wanda  & 78.56 & 73.81 & 75.14 & 43.60 & 75.81 & 86.33 & 77.53 & 58.02 \\
                          & & OATS  & 81.07 & 79.18 & 76.09 & 50.20 & 74.73 & 87.77 & 80.05 & 58.02 \\ \hline %
\multirow{16}{*}{Llama-3 8B} & 0\% & Dense & 80.74 & 79.16 & 73.40 & 45.0 0& 67.87 & 80.98 & 77.69 & 53.50 \\  \cline{2-11}
 & \multirow{5}{*}{30\%} & SparseGPT  & 80.36 & 78.58 & 73.24 & 44.40 & 66.79 & 81.38 & 76.81 & 51.11 \\ 
                          & & Wanda  & 79.98 & 78.00 & 73.64 & 44.40 & 64.26 & 81.62 & 76.18 & 50.94 \\ 
                          & & DSNoT w/ SparseGPT  & 80.20 & 78.12 & 73.80 & 44.40 & 65.70 & 82.20 & 75.72 & 51.71 \\
                          & & DSNoT w/ Wanda  & 79.82 & 77.99 & 73.09 & 44.80 & 63.18 & 81.80 & 77.06 & 51.37 \\
                          & & OATS  & 80.03 & 78.75 & 73.64 & 45.20 & 66.06 & 81.13 & 76.94 & 52.99 \\ \cline{2-11} %
& \multirow{5}{*}{40\%}  & SparseGPT  & 79.16 & 76.74 & 73.32 & 41.80 & 64.26 & 81.31 & 74.71 & 49.32 \\ 
                          & & Wanda  & 78.73 & 75.90 & 72.22 & 44.40 & 63.18 & 80.46 & 72.31 & 49.15 \\ 
                          & & DSNoT w/ SparseGPT  & 78.29 & 75.92 & 73.32 & 42.60 & 58.48 & 80.86 & 73.11 & 47.70 \\
                          & & DSNoT w/ Wanda  & 78.51 & 75.52 & 73.24 & 43.80 & 61.73 & 80.70 & 72.01 & 47.70 \\
                          & & OATS  & 79.71 & 77.18 & 74.19 & 43.80 & 67.51 & 82.39 & 74.92 & 49.74 \\ \cline{2-11} %
& \multirow{5}{*}{50\%}    & SparseGPT  & 77.58 & 73.12 & 72.85 & 40.80 & 59.21 & 79.30 & 69.28 & 45.14 \\ 
                          & & Wanda  & 77.53 & 69.34 & 70.24 & 40.00 & 61.73 & 76.57 & 66.96 & 43.77 \\ 
                          & & DSNoT w/ SparseGPT  & 76.88 & 69.45 & 69.30 & 39.60 & 59.21 & 77.25 & 67.93 & 43.32 \\
                          & & DSNoT w/ Wanda  & 77.09 & 68.57 & 69.77 & 38.60 & 57.76 & 76.27 & 67.34 & 43.43 \\
                          & & OATS  & 77.75 & 73.17 & 71.74 & 41.00 & 64.98 & 79.66 & 72.35 & 45.05 \\ \hline
\multirow{16}{*}{Llama-3 70B} & 0\% & Dense & 84.33 & 84.89 & 80.35 & 48.60 & 68.23 & 85.26 & 86.03 & 64.51 \\  \cline{2-11}
 & \multirow{5}{*}{30\%} & SparseGPT  & 84.66 & 84.63 & 80.35 & 48.00 & 69.31 & 85.26 & 85.02 & 63.31 \\ 
                          & & Wanda  & 84.39 & 83.97 & 80.58 & 48.40 & 70.04 & 85.29 & 85.06 & 63.82 \\ 
                          & & DSNoT w/ SparseGPT  & 84.06 & 84.49 & 80.11 & 48.20 & 69.68 & 85.57 & 85.10 & 63.82 \\
                          & & DSNoT w/ Wanda  & 84.55 & 84.48 & 81.22 & 47.80 & 71.12 & 85.93 & 85.06 & 64.16 \\
                          & & OATS  & 84.28 & 84.40 & 80.66 & 48.40 & 69.31 & 85.32 & 85.90 & 63.65 \\ \cline{2-11} %
& \multirow{5}{*}{40\%}  & SparseGPT  & 83.62 & 83.77 & 80.03 & 47.80 & 69.68 & 85.69 & 84.47 & 61.95 \\ 
                          & & Wanda  & 83.57 & 83.03 & 78.93 & 47.40 & 68.23 & 85.05 & 84.34 & 62.29 \\ 
                          & & DSNoT w/ SparseGPT  & 82.37 & 83.21 & 78.85 & 46.20 & 66.43 & 85.20 & 83.75 & 60.07 \\
                          & & DSNoT w/ Wanda  & 83.79 & 83.35 & 79.72 & 46.80 & 67.87 & 85.57 & 84.89 & 62.37 \\
                          & & OATS  & 84.44 & 83.69 & 80.11 & 48.60 & 70.40 & 84.56 & 84.55 & 62.71 \\ \cline{2-11} %
& \multirow{5}{*}{50\%}    & SparseGPT  & 83.13 & 81.68 & 79.32 & 46.20 & 71.12 & 85.17 & 81.27 & 57.51 \\ 
                          & & Wanda  & 83.08 & 81.12 & 78.22 & 48.00 & 69.31 & 84.22 & 81.61 & 57.25 \\ 
                          & & DSNoT w/ SparseGPT  & 81.34 & 80.68 & 77.82 & 45.60 & 70.04 & 84.62 & 80.98 & 55.12 \\
                          & & DSNoT w/ Wanda  & 85.24 & 81.64 & 78.45 & 46.80 & 69.31 & 85.23 & 81.52 & 57.76 \\
                          & & OATS  & 83.41 & 82.16 & 79.01 & 47.40 & 68.59 & 85.47 & 82.11 & 58.28 \\ 
\bottomrule
\end{tabular}
\end{adjustbox}
\caption{Task-Specific Zero-Shot Results} \label{table:zero_shot_specific}
\end{table}
\subsection{Wikitext-2 Perplexity}\label{app:zero_shot} 
We utilized LM Harness \citep{eval-harness} to measure the Wikitext-2 perplexities for the Phi-3 Mini, Medium, and Llama-3 8B models in Table \ref{table:ppl}. However, prior works \citep{sun2024a, zhang2024dynamic} have also opted to measure Wikitext-2 perplexities utilizing a different normalization and dataset segmentation. We report the Wikitext-2 perplexities utilizing the latter \footnote{We use the implementation sourced from the code base of \citet{sun2024a}: \href{https://github.com/locuslab/wanda/blob/main/lib/eval.py}{https://github.com/locuslab/wanda/blob/main/lib/eval.py} } approach in Table \ref{app:ppl_prior} below:  
\begin{table}[H]
\centering

\begin{tabular}{lcccc}
\toprule
 \multirow{2}{*}{\textbf{Compression}} & \multirow{2}{*}{\textbf{Method}} & \multicolumn{2}{c}{\textbf{Phi-3}} & \multicolumn{1}{c}{\textbf{Llama-3}} \\ 
\cmidrule(lr){3-5}
 &  & \textbf{Mini ($\mathbf{3.8}$B)} & \textbf{Medium ($\mathbf{14}$B)} & \textbf{8B} \\ 
\midrule
0\% & Dense   & 5.64 & 4.02 & 5.54  \\ 
\midrule 
\multirow{4}{*}{30\%} & SparseGPT  & 6.32 & 4.56 & 6.04  \\ 
& Wanda     &  6.19 & 4.46 & 6.03\\ 
& DSNoT     &  6.11 & 4.40 & 5.99 \\ 
& \algname & \textbf{5.96} & \textbf{4.30} & \textbf{5.87}\\
\midrule
\multirow{4}{*}{40\%} & SparseGPT  & 7.05 & 5.02 & 6.75 \\ 
& Wanda     &  6.99 & 4.97 & 6.69 \\ 
& DSNoT     &  6.81 & 4.89 & 6.67 \\ 
& \algname & \textbf{6.51} & \textbf{4.64} & \textbf{6.39} \\
\midrule 
\multirow{4}{*}{50\%} & SparseGPT  & 8.55 & 5.65 & 8.30  \\ 
& Wanda     &  8.84 & 5.73 & 8.79 \\ 
& DSNoT     &  8.58 & 5.68 & 8.84 \\ 
& \algname & \textbf{7.98} & \textbf{5.28} & \textbf{7.77} \\
\bottomrule
\end{tabular}
\caption{Comparison of perplexity (lower is better) utilizing a different normalization and dataset segmentation.} \label{app:ppl_prior}
\end{table}
\subsection{Pruning Implementation and Hyperparameters}\label{app:pruning_hyper}
Our code and pruning method implementations are based on the following codebases:
\begin{itemize}
    \item SliceGPT \citep{ashkboos2024slicegpt}: \href{https://github.com/microsoft/TransformerCompression}{https://github.com/microsoft/TransformerCompression}
    \item SparseGPT \citep{frantar-sparsegpt}: \href{https://github.com/IST-DASLab/sparsegpt}{https://github.com/IST-DASLab/sparsegpt}
    \item Wanda \citep{sun2024a}: \href{https://github.com/locuslab/wanda}{https://github.com/locuslab/wanda}
    \item DSNoT \citep{zhang2024dynamic}: \href{https://github.com/zyxxmu/DSnoT}{https://github.com/zyxxmu/DSnoT}
    \item OWL \citep{yin2024outlier}: \href{https://github.com/luuyin/OWL}{https://github.com/luuyin/OWL}
\end{itemize}
We utilize Huggingface's Transformers library to implement the large language models and vision transformers for our experiments \citep{wolf-etal-2020-transformers}. 
\subsubsection{SparseGPT Hyperparameters}
We utilize a blocksize of 128 across all experiments and a Hessian dampening of $0.01$ and $0.1$ where the ladder is utilized only when faced with non-positive definiteness issues related with the Cholesky decomposition. 
\subsubsection{DSNoT Hyperparameters}
We run experiments utilizing DSNoT where the initial masks are generated by SparseGPT and Wanda. All DSNoT experiments were run with $50$ iterations and an update threshold of $0.1$. Table \ref{table:dsnot_specific}, below, shows the results distinguishing between the two initial methods that were utilized.  
\begin{table}[H]
\centering
\begin{tabular}{l|ccccc}
\toprule
\textbf{Model} & \textbf{Compression} & \textbf{Method}  & \textbf{MMLU ($\uparrow$)} & \textbf{Zero-Shot($\uparrow$)} & \textbf{Perplexity($\downarrow$)} \\ \hline
\multirow{6}{*}{Phi-3 Mini} & \multirow{2}{*}{30\%}  & DSNoT w/ SparseGPT  &67.01 & 70.81 & 10.55 \\ 
                          & & DSNoT w/ Wanda  & 68.02 & 71.20 & 10.51  \\ \cline{2-6}
& \multirow{2}{*}{40\%}  & DSNoT w/ SparseGPT\ & 62.94 & 68.86 & 12.29 \\
                        & & DSNoT w/ Wanda  & 63.57 & 69.08 & 12.17 \\ \cline{2-6}
& \multirow{2}{*}{50\%}    & DSNoT w/ SparseGPT  & 53.99 & 64.74 & 16.71  \\ 
                          & & DSNoT w/ Wanda  & 54.28 & 65.33 & 16.68  \\ \hline
\multirow{6}{*}{Phi-3 Medium} & \multirow{2}{*}{30\%}  & DSNoT w/ SparseGPT  &74.89 & 73.82 & 7.11 \\ 
                          & & DSNoT w/ Wanda  & 75.13 & 74.03 & 7.11  \\ \cline{2-6}
& \multirow{2}{*}{40\%}  & DSNoT w/ SparseGPT\ & 73.15 & 72.54 & 8.24 \\
                        & & DSNoT w/ Wanda  & 73.20 & 72.90 & 8.27 \\ \cline{2-6}
& \multirow{2}{*}{50\%}    & DSNoT w/ SparseGPT  & 68.65 & 71.12 & 9.96  \\ 
                          & & DSNoT w/ Wanda  & 68.12 & 71.10 & 10.02  \\ \hline
\multirow{6}{*}{Llama-3 8B} & \multirow{2}{*}{30\%}  & DSNoT w/ SparseGPT  &62.99 & 68.98 & 9.37 \\ 
                          & & DSNoT w/ Wanda  & 63.72 & 68.64 & 9.36  \\ \cline{2-6}
& \multirow{2}{*}{40\%}  & DSNoT w/ SparseGPT\ & 58.97 & 66.28 & 9.60 \\
                        & & DSNoT w/ Wanda  & 59.99 & 66.65 & 9.68 \\ \cline{2-6}
& \multirow{2}{*}{50\%}    & DSNoT w/ SparseGPT  & 49.15 & 62.74 & 12.41  \\ 
                          & & DSNoT w/ Wanda  & 49.20 & 62.35 & 12.42  \\\hline
\multirow{6}{*}{Llama-3 70B} & \multirow{2}{*}{30\%}  & DSNoT w/ SparseGPT  &78.76 & 75.13 & 3.28 \\ 
                          & & DSNoT w/ Wanda  & 79.00 & 75.54 & 3.27  \\ \cline{2-6}
& \multirow{2}{*}{40\%}  & DSNoT w/ SparseGPT\ & 76.39 & 73.26 & 4.16 \\
                        & & DSNoT w/ Wanda  & 77.70 & 74.29 & 4.10 \\ \cline{2-6}
& \multirow{2}{*}{50\%}    & DSNoT w/ SparseGPT  & 72.18 & 72.02 & 5.87  \\ 
                          & & DSNoT w/ Wanda  & 72.76 & 72.91 & 5.58  \\
\bottomrule
\end{tabular}
\caption{LLM performance metrics of DSNoT with different initial methods.} \label{table:dsnot_specific}
\end{table}
Table \ref{table:dsnot_specific_vit}, below, shows the analogous results but for our vision transformer experiments: 
\begin{table}[H]
\centering
\begin{tabular}{l|ccc}
\toprule
\textbf{Model} & \textbf{Compression} & \textbf{Method}  & \textbf{Accuracy (\%)} \\ \hline
\multirow{6}{*}{ViT-Base} & \multirow{2}{*}{30\%}  & DSNoT w/ SparseGPT  &80.01  \\ 
                          & & DSNoT w/ Wanda  & 80.16  \\ \cline{2-4}
& \multirow{2}{*}{40\%}  & DSNoT w/ SparseGPT\ & 79.12  \\
                        & & DSNoT w/ Wanda  & 79.46 \\ \cline{2-4}
& \multirow{2}{*}{50\%}    & DSNoT w/ SparseGPT  & 75.83\\ 
                          & & DSNoT w/ Wanda  & 76.90  \\ \hline
\multirow{6}{*}{DinoV2-Giant} & \multirow{2}{*}{30\%}  & DSNoT w/ SparseGPT  &86.46 \\ 
                          & & DSNoT w/ Wanda  & 86.45  \\ \cline{2-4}
& \multirow{2}{*}{40\%}  & DSNoT w/ SparseGPT\ & 86.37  \\
                        & & DSNoT w/ Wanda  & 86.30 \\ \cline{2-4}
& \multirow{2}{*}{50\%}    & DSNoT w/ SparseGPT  & 85.87 \\ 
                          & & DSNoT w/ Wanda  & 85.93  \\
\bottomrule
\end{tabular}
\caption{ImageNet Validation Accuracy of DSNoT with different initial methods.} \label{table:dsnot_specific_vit}
\end{table}
\end{document}